\documentclass{article} 
\usepackage[main, final]{neurips_2026}
\usepackage{hyperref}
\usepackage{xcolor}
\usepackage{float}
\usepackage{amsmath,amssymb,amsfonts}
\usepackage{graphicx}
\usepackage{textcomp}
\usepackage{multirow}
\usepackage{subcaption} 
\usepackage{booktabs} 
\usepackage{wrapfig}
\usepackage{graphicx}
\usepackage{nicefrac}       
\usepackage{microtype}      

\usepackage{acro}

\DeclareAcronym{BP}{
  short = BP ,
  long  = backpropagation ,
  short-plural = s ,
  long-plural  = s
}

\DeclareAcronym{FA}{
  short = FA ,
  long  = feedback alignment ,
  short-plural = s ,
  long-plural  = s
}
\DeclareAcronym{LDFA}{
  short = LDFA ,
  long  = low-dimensional feedback alignment ,
  short-plural = s ,
  long-plural  = s
}

\DeclareAcronym{DFA}{
  short = DFA ,
  long  = direct feedback alignment ,
  short-plural = s ,
  long-plural  = s
}

\DeclareAcronym{SGD}{
  short = SGDm ,
  long  = SGD with momentum ,
  short-plural = s ,
  long-plural  = s
}

\DeclareAcronym{RFLO}{
  short = RFLO ,
  long  = random feedback local online learning ,
  short-plural = s ,
  long-plural  = s
}

\DeclareAcronym{RNN}{
  short = RNN ,
  long  = recurrent neural network ,
  short-plural = s ,
  long-plural  = s
}

\DeclareAcronym{CNN}{
  short = CNN ,
  long  = convolutional neural network ,
  short-plural = s ,
  long-plural  = s
}

\DeclareAcronym{vae}{
  short = VAE ,
  long  = variational autoencoder ,
  short-plural = s ,
  long-plural  = s
}

\DeclareAcronym{BN}{
  short = BN ,
  long  = Batch Normalisation ,
  short-plural = s ,
  long-plural  = s
}
\DeclareAcronym{LN}{
  short = LN ,
  long  = Layer Normalisation ,
  short-plural = s ,
  long-plural  = s
}

\newcommand{\figref}[1]{Figure~\ref{#1}}

\date{}

\title{Overcoming Rank Collapse in Feedback Alignment }
\author{
  Gauthier Boeshertz \\
  Department of Bioengineering \\
  Imperial College London \\ 
  \And
  Razvan Pascanu\thanks{Co-senior Authors} \\
  Mila \\
  \And
  Claudia Clopath\footnotemark[1]  \\
  Department of Bioengineering\\
  Imperial College London \\
}

\begin{document}

\maketitle
\newcommand{\abspercif}{45.7}

\begin{abstract}

\Ac{BP} is widely viewed as biologically implausible, in part because it requires feedback weights to be the transpose of forward weights for error propagation. 
Interestingly, when training a network with fixed random feedback weights to circumvent this issue, learning aligns the forward weights with the feedback weights, leading the backpropagated error signal to become an approximation of the standard gradient used by \ac{BP}.
This process, called \ac{FA}, occurs in MLPs and very shallow CNNs but does not scale well to deeper architectures. In this work, we first investigated differences between \ac{BP} and different \ac{FA} methods, trained on CIFAR10, specifically focusing on the effective rank of the error signal.
We found that \ac{FA} methods produce pseudo-gradient signals that have a considerably lower rank and hence are constrained to a lower-dimensional subspace compared to \ac{BP}, limiting exploration of the parameter space. 
Motivated by this observation, we evaluated two mechanisms for increasing the effective dimensionality of FA: Muon, an optimiser that orthogonalises weight updates, and activity normalisation, which promotes activation orthogonality.
Across larger architectures and benchmarks, we find that these methods consistently improve over \ac{FA} baselines, for example, on ImageNet with AlexNet, adding \ac{BN} increases \ac{FA} accuracy from 0.1\% to 11.3\%, using Muon on top of \ac{BN} increases the accuracy to 20.2\%. 
Our results identify low-dimensional gradient dynamics as a key limitation of \ac{FA}, and suggest that inducing higher-dimensional update geometry is a promising route toward scaling alternatives to \ac{BP}.

\end{abstract}

\acresetall

\section{Introduction}
\label{sec:intro}

Most neural networks are trained using error \ac{BP} \citep{rumelhart1986learning}. This method performs credit assignment by computing the gradient of a loss function for every parameter inside the network. 
A biological view of this process says that the neurons project an error signal backwards, using synapses that share the same weights as the synapses projecting upwards. Sharing the weight between the forward and backward connections has been one of the main issues with regard to the biological plausibility of \ac{BP} \citep{crick1989recent,lillicrap2016random,lillicrap2020backpropagation}. 
However, \citep{lillicrap2016random} has shown that it is possible to train networks with random feedback projections and achieve results similar to \ac{BP} on simple tasks. They observed that throughout training, the forward weights would align to the random feedback, allowing a meaningful error to be backpropagated, as quantified by the alignment of the gradients with the random feedback and gradients obtained with weight sharing. While the error signal propagated in \ac{FA} is not a gradient of the loss, we will refer to it as a gradient to be consistent with prior works.

This observation created a vast literature on making more biological training methods using random feedback projections. \Ac{DFA} \citep{nokland2016direct} showed that it is possible to directly feed the error signal from the output to any hidden layer using feedback projections of the appropriate size. Building on this idea, \citep{frenkel2021learning} showed that even the error can be replaced by the target itself. 

However all of these methods fail to scale to more difficult tasks \citep{liao2016important,xiao2018biologically, moskovitz2018feedback, launay2019principled,bartunov2018assessing}. 
For example, \citep{launay2019principled} reported that \acp{CNN} trained with \ac{DFA} do not perform better than chance on CIFAR100. In general, \ac{FA} methods have been found to perform much worse than \ac{BP} even in shallow \acp{CNN}. 
The reason behind this difference in performance was attributed to the lack of flexibility of convolutional layers due to weight sharing \citep{refinetti2021align,launay2019principled}. To mitigate this lack of flexibility, several methods use the signs of the forward weights to set the signs of the feedback \citep{liao2016important,xiao2018biologically,moskovitz2018feedback}, and achieved performance closer to that of \ac{BP}.
 Instead of imposing a shared structure, some methods start with random feedback and train the feedback along with the forward weights \citep{akrout2019deep,kunin2020two,hanut2025training}, and reach performance even closer to \ac{BP}. 

In this work, we focus on three different formulations of \ac{FA}. The original \ac{FA} \citep{lillicrap2016random}, \ac{DFA} \citep{nokland2016direct}, and LDFA \citep{hanut2025training}, which trains low rank feedback weights. 
Our objective is not to completely close the gap between \ac{BP} and \ac{FA}. Rather, it is to understand \ac{FA} in deep networks and explain what stops alignment from happening. To pursue this objective, we investigated the differences in the geometry of the gradients in networks trained using \ac{BP} and \ac{FA}. We find that in deep \ac{FA} networks, when the weights start aligning, there is a marked decrease in the effective rank \citep{roy2007effective} of gradients. 
This means that the gradients will have a few directions that dominate the others, which may not allow the alignment process to continue, since the parameter updates have to explore much smaller subspaces. 
We therefore try two different methods aimed at increasing the dimensionality of the gradients. When writing dimensionality, we mean the effective dimensionality or rank. 

First, we used a recently proposed optimiser, Muon \citep{bernstein2024old,bernstein2025deriving,jordan2024muon}. Muon differs from standard SGD with momentum by adding an orthogonalisation step to the momentum update. This means setting every singular value to 1. 
The second way is to normalise the activity of the layers using \ac{BN}. \ac{BN} was shown to orthogonalise activity in finite width random networks \citep{daneshmand2020batch,daneshmand2021batch}. \citet{liao2016important} had already showed that \ac{BN} had a positive effect on the performance of \ac{FA} models, but had not explained why. We hypothesise that this is due to the decorrelation of the activations that occurs when using \ac{BN}. \citet{offergeld2026correlationsruininggradientdescent} showed that a method aimed at decorrelating activity improved learning in \ac{BP} and \ac{FA} networks. 
Overall, we analyse multiple feedback methods and investigate the effects of the different interventions by studying how low-dimensional gradient dynamics can limit alignment in deep networks. Through our attempt of understanding \ac{FA}, we also provide new perspectives on the effects of widely used optimisation techniques that differ from those usually obtained in \ac{BP}.

We tested these methods on the CIFAR10 \citep{krizhevsky2009learning} dataset and found that they improved the performance of \ac{FA} models considerably. Using them, the effective dimensionality of the gradients did not exhibit the marked decrease observed before, which supports the claim that \ac{FA} requires gradients to have large rank for alignment to happen. 
We then measured the impact of the two methods using ResNet-18 \citep{he2016deep} and AlexNet \citep{krizhevsky2012imagenet} on STL-10 \citep{coates2011analysis}, CIFAR-100 \citep{krizhevsky2009learning}, and ImageNet \citep{russakovsky2015imagenet}. Both interventions improved \ac{FA} and \ac{DFA}, with their combination giving the highest mean accuracy in every reported benchmark.

\section{Optimisation with Feedback Alignment}

\subsection{Backpropagation and Feedback Alignment}

\begin{figure}
    \centering
    \includegraphics[width=0.95\linewidth]{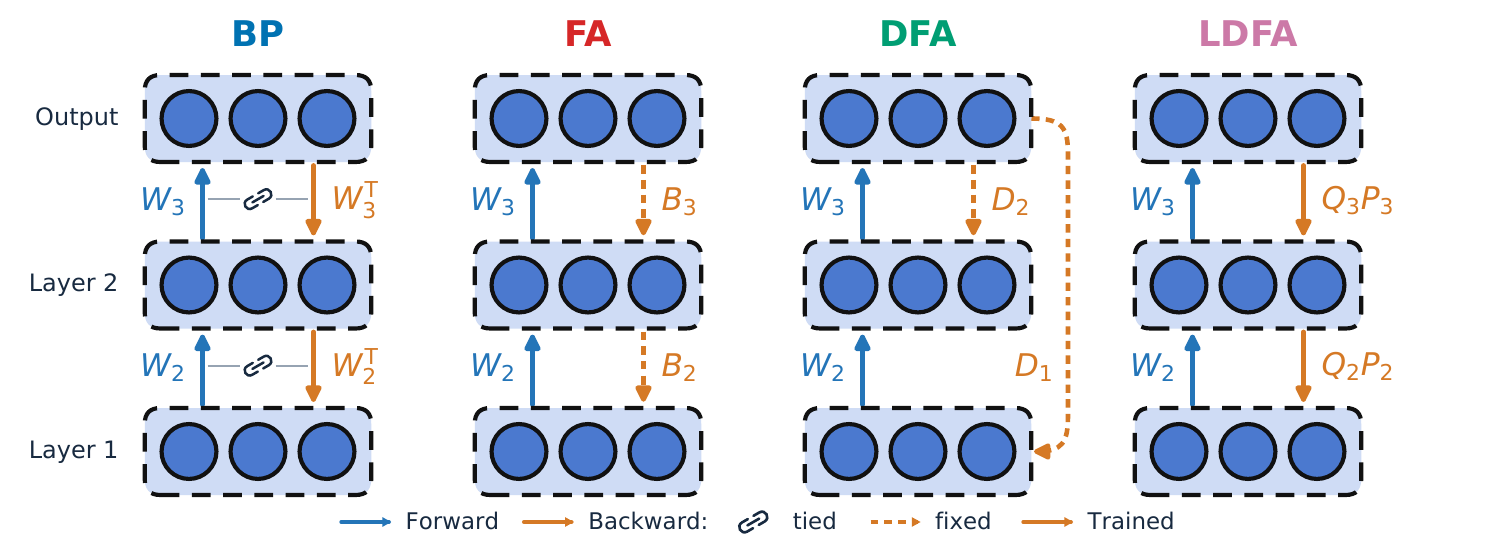}
    \caption{\textbf{Feedback pathways for the four learning rules.} \Ac{BP} uses transposed forward weights; \ac{FA} uses fixed random connections between adjacent layers; \ac{DFA} uses fixed projections of the output error directly to hidden layers; and LDFA learns the factors $Q_\ell,P_\ell$ of each layerwise feedback map.  Blue arrows denote forward connections and orange arrows denote feedback.}
    \label{fig:feedback_methods}
\end{figure}

\Ac{BP} \citep{rumelhart1986learning} computes gradients by traversing the computation graph from the output layer back to the input, propagating error signals layer by layer through the transpose of the forward weights. For a standard MLP network with \(L\) layers:
\begin{equation}
a_\ell = W_\ell h_{\ell-1} + b_\ell,
\quad
h_\ell = \phi(a_\ell)
\end{equation}
where $a_\ell$ is the pre-activation vector, $h_\ell$ is the post-activation vector, $W_\ell$ is the weight matrix, $b_\ell$ is the bias vector, and $\phi(\cdot)$ is the activation function, for layer  \(\ell=1,\dots,L\). Using the loss computed at the output of the network, \ac{BP} computes the error for each layer recursively:
\begin{equation}
\delta_L = \frac{\partial E}{\partial a_L},
\quad
\delta_{\ell}^{BP} = \left(W_{\ell+1}^\top \delta_{\ell+1}^{BP} \right)\odot \phi'(a_{\ell})
\quad \text{for } \ell = L-1,\dots,1
\label{eq:bp_rec}
\end{equation}

where $\odot$ denotes the Hadamard product, $\phi'$ is the derivative of the activation function. The use of the transposed weight in Eq. \ref{eq:bp_rec} is the core of the Weight Transport problem. Training a network with \ac{FA} replaces the transpose with a weight $B_{\ell}^{FA}$, usually sampled from the same distribution as the forward weights, and keeps it fixed during training. \ac{DFA} projects the output error directly to the hidden layers using feedback weights $B_{\ell}^{DFA}$:

\begin{equation}
\delta_{\ell}^{FA} = \left(B_{\ell+1}^{FA} \delta_{\ell+1}^{FA} \right)\odot \phi'(a_{\ell}),
\quad \delta_{\ell}^{DFA} = \left(B_{\ell}^{DFA} \delta_{L} \right)\odot \phi'(a_{\ell})
\quad \text{for } \ell = L-1,\dots,1
\label{eq:fa_rec}
\end{equation}

To investigate the impact of low-dimensional errors, we also considered \ac{LDFA} \citep{hanut2025training}, which uses learned, factorised feedback maps $B_\ell^{LDFA}=Q_\ell P_\ell$. Writing $n_\ell$ for the width of layer $\ell$, the factors have dimensions $P_\ell\in\mathbb{R}^{r_\ell\times n_\ell}$ and $Q_\ell\in\mathbb{R}^{n_{\ell-1}\times r_\ell}$, so $B_\ell^{LDFA}$ has rank at most $r_\ell$. The projection $P_\ell$ compresses the incoming error to $r_\ell$ components, and $Q_\ell$ maps these components back to the preceding layer:
\begin{equation}
\delta_{\ell}^{LDFA} = \left(Q_{\ell+1}P_{\ell+1}\delta_{\ell+1}^{LDFA}\right)\odot\phi'(a_\ell)
\quad \text{for } \ell = L-1,\dots,1.
\label{eq:ldfa_rec}
\end{equation}

All four methods use the same output error $\delta_L$. The errors are multiplied by the activity of the layer the weights project from to make the gradient, or pseudo-gradient for the \ac{FA} methods:
\begin{equation}
\frac{\partial E}{\partial W_\ell} = \delta_\ell^{(BP,FA,DFA,LDFA)} (h_{\ell-1})^\top,
\quad
\frac{\partial E}{\partial b_\ell} = \delta_\ell^{(BP,FA,DFA,LDFA)}
\end{equation}
where we set $h_0$ to be the input to the network.

Although we wrote the equations for an MLP, the same ideas apply to \acp{CNN}. In convolutional layers, backpropagation still propagates errors backward and forms gradients from error signals and layer activations, but the linear maps are convolutional operators with shared weights across spatial locations rather than dense matrices. Accordingly, the backward pass uses the adjoint convolution operator, and \ac{FA} replaces this operator with a fixed random feedback map of compatible shape. For \ac{LDFA}, a learned $1\times1$ projection compresses the error channels before applying the adjoint convolution associated with the learned spatial kernel $Q_\ell$. 

Additional implementation details for all four methods are provided in Appendix~\ref{sec:feedback_implementation}.

\subsection{Optimisation}
Once gradients have been computed, their use is determined by the optimiser, which is separate from the choice of learning rule used to obtain them. \ac{FA} can therefore use any optimiser that works for \ac{BP}. We will especially be interested in a recently proposed optimiser, called Muon \citep{bernstein2024old,bernstein2025deriving,jordan2024muon}. This optimiser first received attention because it leads to faster convergence under mild additional computational costs  \citep{jordan2024muon}, and recently because it leads to lower loss for the same compute budget in large language models \citep{liu2025muon}. It works similarly to SGD with momentum, where $\beta$ is the momentum parameter, but adds an orthogonalisation step: 
\begin{equation}
M_t = \beta M_{t-1} + (1-\beta)\frac{\partial E}{\partial W^\ell},
\quad 
W_{t+1}^{\ell}
=
W_t^{\ell}
-
\eta\,  \operatorname{Orth}(M_t^{\ell})
\quad
\end{equation}
 The orthogonalisation function $\operatorname{Orth}$ computes the closest orthogonal matrix to its input, which can be written in terms of the SVD of the input, $X = U \Sigma V^\top$, then the orthogonalised matrix is given by $\operatorname{Orth}(X) = U V^\top.$ This operation is usually done with the Newton-Schulz approximation \citep{jordan2024muon}, which is quicker than computing the singular vectors.  An intuitive explanation of the orthogonalisation is given in \citep{jordan2024muon}, which we rely on for our results. It states that updates in standard optimisers have low rank, and orthogonalising the update will make the lesser explored directions more relevant. For vector parameters which cannot have their updates orthogonalised and need to use AdamW \citep{loshchilov2017decoupled}. We use a version of Muon that scales the updates to have a similar RMS norm to AdamW \citep{liu2025muon}, which allows the reuse of the learning rate across parameters trained with AdamW and Muon.
\section{Pathologies of \ac{FA}}
\label{sec:patho}

\begin{figure}[!htbp]
    \centering
    \subfloat{\includegraphics[width=0.95\linewidth]{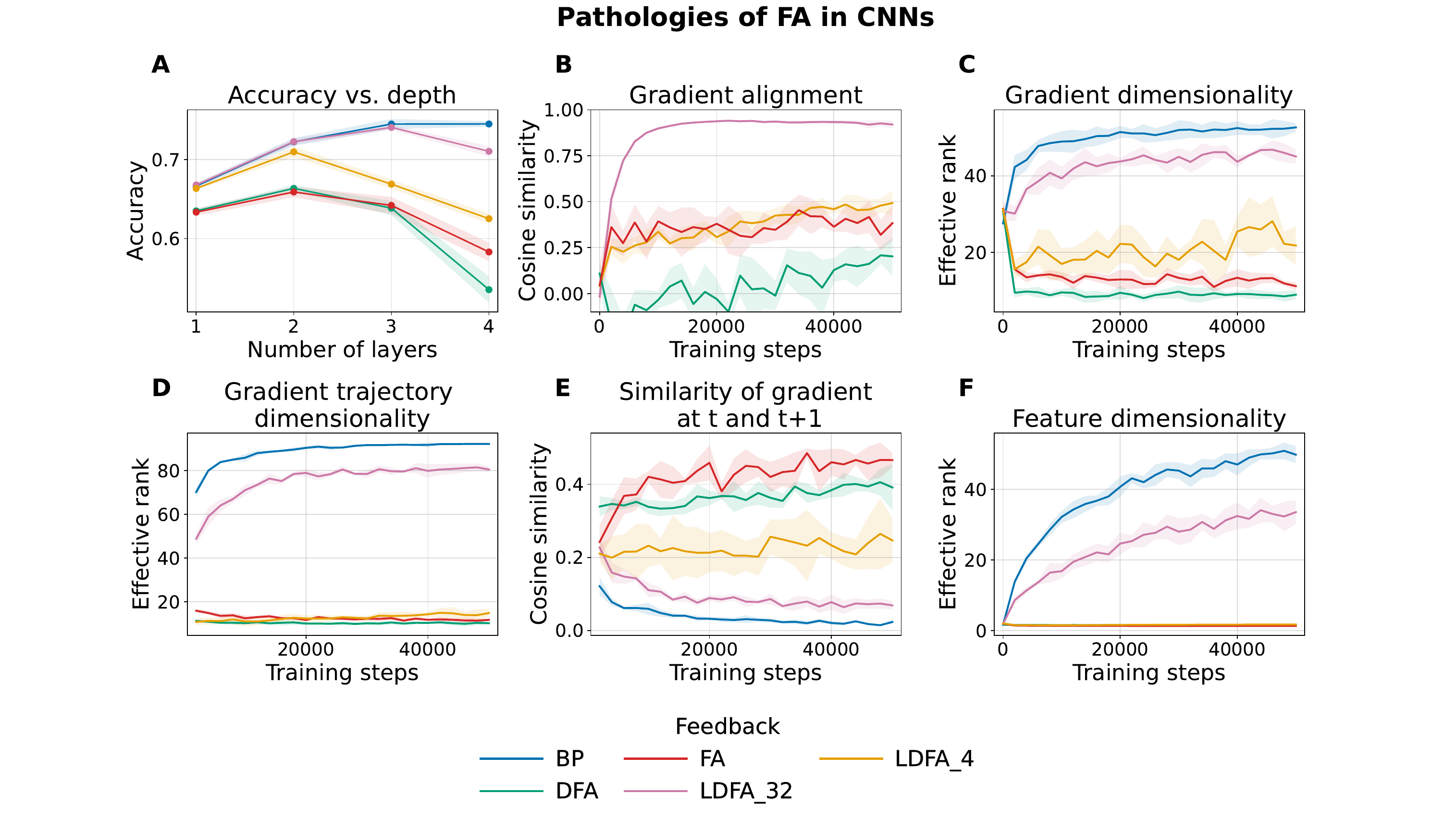}}
    \caption{\textbf{Deep \ac{FA} networks suffer from low dimensional gradients.} \textbf{A} Increasing the number of layers results in lower performance in \ac{FA} networks. \textbf{B-F} Metrics for the last layer of the 4 convolutional layer model. \textbf{B} Gradient alignment measured by the cosine similarity of the gradient obtained using the feedback weight and the transpose of the forward weights for the feedback projections. \textbf{C} Effective rank of the gradient in the last convolutional layer. The \ac{FA} networks have much lower dimensionality. \textbf{D} Effective rank of the Gram matrix of the gradients. Again, \ac{FA} networks have much lower dimensionality.  \textbf{E} Cosine similarity of the gradients at step t and t+1. \ac{FA} networks have similar gradients, which explains the low-dimensional gradient trajectory. \textbf{F} The feature dimensionality is much lower than for the \ac{BP} models.}
    \label{fig:fig1}
\end{figure}

We first explored the effects of \ac{FA} in deep networks to understand why it performs substantially worse than \ac{BP} in deep \acp{CNN}. We trained \acp{CNN} with depth varying from 1 to 4 before the classifier. Each convolutional layer had 64 channels, a 3x3 kernel, and was followed by 2x2 max-pooling; on the CIFAR10 \citep{krizhevsky2009learning} dataset. 
Using more channels leads to a decrease of the gap between \ac{FA} and \ac{BP} \citep{boopathy2022train}; however, in sufficiently deep networks, this gap remains large, leading us to focus on the settings in which \ac{FA} fails.
 We use 64 as it strikes a balance between too narrow and too wide networks, and because it usually is the number of channels in the first layer of deep CNNs, for example, Resnets \citep{he2016deep} and VGGs \cite{simonyan2014very}. 
 We used ReLU activations, as it is the activation used in these deep networks. For the optimiser, we used AdamW \citep{loshchilov2017decoupled}, but note that SGD with momentum gives similar results. The weights of the convolutional and linear layers were initialised using the Kaiming uniform initialisation as in \citep{frenkel2021learning}. 
 We also found that it was the one that worked best for \ac{FA} networks, which heavily depended on having the right initialisation. We first measured the effect of depth on accuracy, using identical architectures for every feedback method at each depth, allowing a matched comparison as convolutional blocks are added. We then turned to diagnostics metrics to investigate the source of the gap (\figref{fig:fig1}).

Consistent with earlier studies \citep{cheon2024pretraining,launay2019principled}, we found that the accuracy of the \ac{FA} models decreases with depth (\figref{fig:fig1} A). To understand this failure of \ac{FA}, we turned our attention to networks with 4 convolutional layers. The gradient alignment, measured by the cosine similarity of the flattened pseudo gradients with the true gradients, as computed by \ac{BP}, increases at first, then stops (\figref{fig:fig1} B). 
 Since the gradient alignment is small, the \ac{FA} gradient will not follow the true gradient of the loss, which explains why the accuracy changes slowly. To explain why \ac{FA} gradient alignment does not increase much after an initial bump, we analysed the geometry of the updates of the different \ac{FA} models.

Guided by the intuition that \ac{FA} makes optimisation happen in lower-dimensional subspaces, we measured three metrics that may explain the failure of \ac{FA} and \ac{DFA} to align to \ac{BP}. They are all based on the effective rank of different matrices. 
The effective rank \citep{roy2007effective} is a continuous approximation to the rank of the matrix. 
It is computed by taking the exponentiated entropy of the normalised singular values of the matrix: $\text{erank}(M)=\exp\!\left(-\sum_{i=1}^r p_i \log p_i \right)$, with $p_i = \frac{\sigma_i}{\sum_{j=1}^r \sigma_j}$ where $\sigma_i$ are the singular values. 
We measured this metric on three different matrices:
\begin{itemize}
    \item On the gradients of the convolutional layers, to measure the dimensionality of the error propagated by different methods (\figref{fig:fig1} C). 
    Convolutional layers weights have gradients with shape \(C_{\mathrm{out}}\times C_{\mathrm{in}}\times k_h\times k_w\), we retain output channels as rows and flatten input channels and kernel coordinates into columns, obtaining \(D_t\in\mathbb{R}^{C_{\mathrm{out}}\times(C_{\mathrm{in}}k_hk_w)}\), as done in the Muon optimizer \citep{jordan2024muon}. We report \(r_{\mathrm{grad}}(t)=\operatorname{erank}(D_t)\).
    \item On the trajectory of the gradients (\figref{fig:fig1} D). We flattened the gradients from the 100 optimisation steps before computing the metric, and computed the effective rank of the Gram matrix of the stacked gradients. Stacking $T=100$ such vectors as rows gives \(H\in\mathbb{R}^{T\times P}\), where \(P\) is the number of weight parameters in that layer. We normalise each row of $H$ to unit $\ell_2$ norm, $\tilde H_{t}=H_{t}/\lVert H_{t}\rVert_2$, so that the metric captures the directions of the gradients and not their magnitudes. The Gram matrix $\tilde H\tilde H^\top$ is then the matrix of cosine similarities between gradients at different steps. We report \(r_{\mathrm{traj}}=\operatorname{erank}(\tilde H\tilde H^\top)\), which is bounded above by \(\min(T,P)\).
    \item On the features of the last convolutional layer (\figref{fig:fig1} F). We computed the Gram matrix using the activations that feed into the classifier, and then computed the effective rank of this matrix, as done in \citep{huh2021low}.
\end{itemize}

We found that the dimensionality of the gradients of \ac{FA} networks is always lower than that of similar \ac{BP} networks (\figref{fig:fig1} C). Moreover, the trajectory of the gradients is also low-dimensional (\figref{fig:fig1} D). This indicates that the gradients are more aligned with one another. In other words, the gradients span a lower-dimensional subspace, so the parameter updates are confined to fewer directions. The optimisation explores only a limited region of parameter space, which may be insufficient to produce meaningful alignment. In agreement with our hypothesis, the dimensionality of the trajectory of \ac{FA} networks decreases early, and stays much lower than that of the \ac{BP} networks (\figref{fig:fig1} D). 
Indeed, while the \ac{BP} trajectory dimensionality almost reaches its maximum rank, 100, the \ac{FA} one is stuck under 20, and ends at 12.  To better explain this, we looked at the entries next to the diagonal of the trajectory Gram matrix, which are the cosine similarity of gradients adjacent in the trajectory. We computed the mean of theses values across the whole matrix and found high cosine similarity for the \ac{FA} network where it is almost zero for the \ac{BP} networks (\figref{fig:fig1} E). 
The trajectory of the gradients will be higher for \ac{BP} as it uses updated weights to propagate the error, which may increase the dimensionality when keeping other parts of the gradient equal. \ac{FA} by construction can not do this, however, it is not the only reason for the lower trajectory dimensionality, as we show in the next section.

One of the effects of low-dimensional gradients could be that the representations will be low-dimensional too. As expected, the features of the \ac{FA} networks are much lower-dimensional than those of \ac{BP} networks, which may be due to the small subspace spanned by the gradients. While low-dimensional features have been found to be beneficial for generalisation \citep{valle2018deep,huh2021low,andriushchenko2023sharpness}, we argue here that the small feature dimensionality may result from limited exploration of the parameter space, which limits the alignment of the weights, and therefore the ability of the network to learn useful task related features.

Together, these results suggest that the poor performance of \ac{FA} in deeper \acp{CNN} is due to imperfect alignment with \ac{BP}, which may be limited because of the geometry of the updates. In particular, \ac{FA} appears to restrict learning to a low-dimensional subspace, producing highly correlated gradients and low-dimensional features that may be insufficient for discriminating between classes. 

\section{Higher-dimensional updates enable better \ac{FA}}
\label{sec:explo}
\begin{figure}
    \centering
    \subfloat{\includegraphics[width=0.95\linewidth]{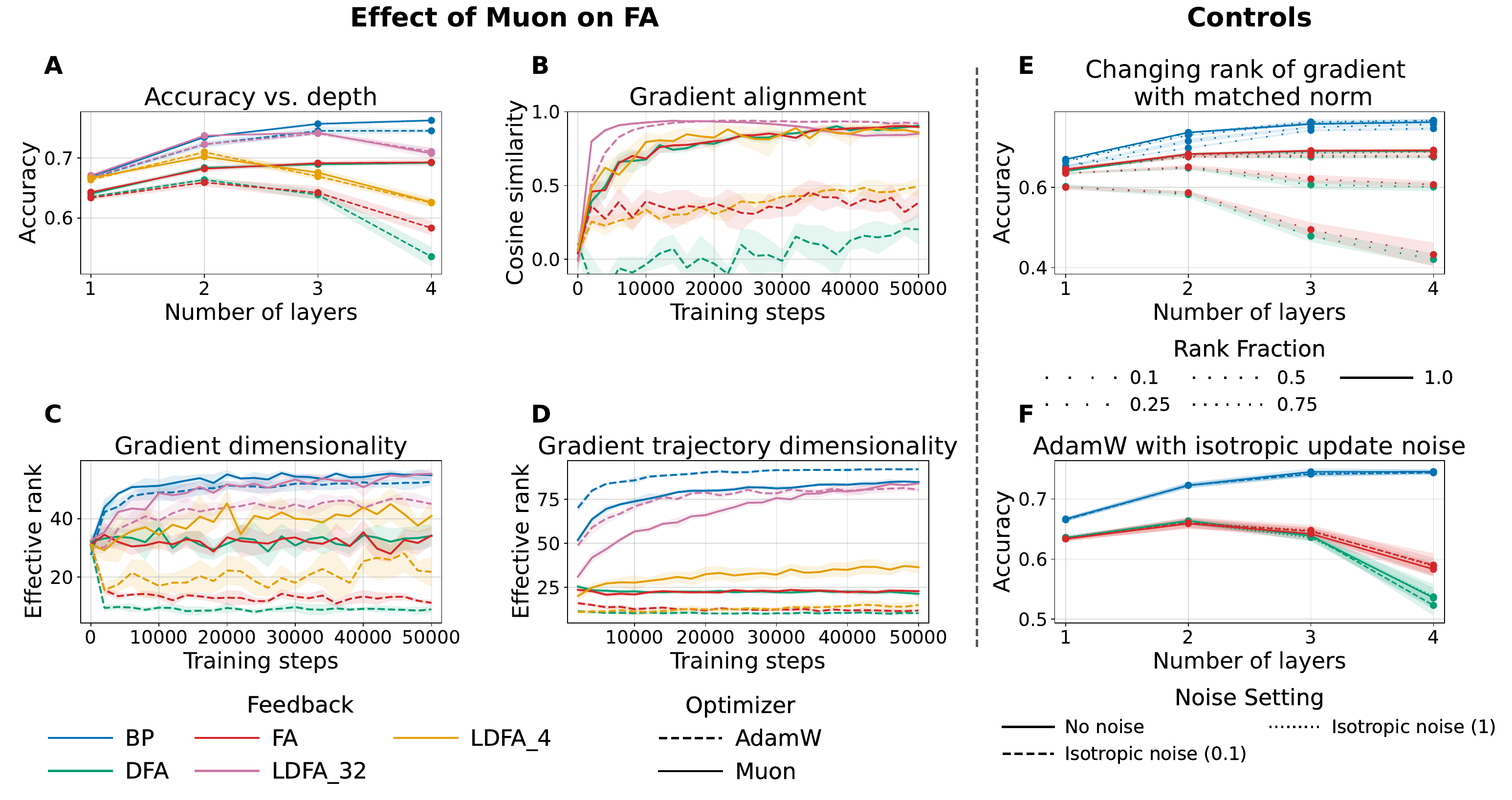}}
    \caption{\textbf{Orthogonalised updates improve \ac{FA} learning, and restricting their rank reduces the benefit.} \textbf{A} Test accuracy versus convolutional depth for AdamW (dashed) and Muon (solid). Muon substantially reduces the loss of accuracy with depth for \ac{FA} and \ac{DFA}. \textbf{B-D} Training dynamics in the last convolutional layer of the four-layer networks.(\textbf{B}) Muon increases the gradient alignment. (\textbf{C}) The effective rank of the gradient matrix. (\textbf{D}) The effective rank of the Gram matrix of the gradient trajectory. The effect on these metrics varies across feedback rules. \textbf{E,F}, controls of possible Muon confounding factors. \textbf{E} Norm-matched Muon retains the leading 10\%, 25\%, 50\%, 75\%, or 100\% of singular directions and equalises their singular values, with the same target update RMS before learning-rate scaling. Retaining more directions improves \ac{FA} and \ac{DFA} accuracy, especially at greater depth. \textbf{F} AdamW with no added noise or isotropic Gaussian noise with norm equal to 0.1 or 1 times the base update norm. Adding noise does not reproduce Muon's accuracy gains in \ac{FA} and \ac{DFA}.}
    \label{fig:fig_muon}
\end{figure}

The previous section suggests a direct intervention: if the failure mode of \ac{FA} is caused by updates that are too low-dimensional, then methods that increase the dimensionality of the updates or of the intermediate representations should improve \ac{FA} training. Therefore, we tried different ways of increasing the dimensionality of the updates, using the same setup as in the previous sections. 

\paragraph{Using orthogonalised updates}
First, we used the Muon optimiser \citep{jordan2024muon}. The low effective rank of the \ac{FA} gradient suggests that a few singular directions dominate the learning signal, while other directions contribute little to the update. Muon flattens the singular-value spectrum of the momentum update, giving greater relative weight to these weaker directions. Since the feedback rule determines how gradients are computed and the optimiser determines how they are applied, we can simply swap AdamW with Muon for each feedback rule.

Compared with AdamW, Muon substantially reduces the degradation in accuracy with depth for both \ac{FA} and \ac{DFA} (\figref{fig:fig_muon} A). In the four-convolutional-layer \ac{FA} network, test accuracy reaches 69.8\% with Muon, compared with 58.3\% with AdamW. The benefits are much smaller for \ac{BP} and \ac{LDFA}. \ac{BP} is already well optimised with AdamW, and in \ac{LDFA}, Muon has less directions to increase dimensionality in due to the nature of the feedback rule. For the four-layer networks, we measured gradient alignment with \ac{BP} (\figref{fig:fig_muon} B), gradient effective rank (\figref{fig:fig_muon} C), and gradient-trajectory effective rank (\figref{fig:fig_muon} D). For \ac{FA} and \ac{DFA}, Muon improves gradient alignment and increases both gradient and trajectory effective rank relative to AdamW. 

To test the contribution of update rank while controlling its scale, we used a norm-matched variant of Muon (\figref{fig:fig_muon} E). We replace the orthogonalisation step with a truncated SVD, retaining only the leading singular directions and normalising the resulting matrix to a fixed target RMS: $ 0.2 \sqrt{\frac{mn}{r}} U_{[:,1:r]}V_{[:,1:r]}^\top$.
We swept over a range of fraction of the total number of singular values to retain. Retaining only a small fraction of directions strongly reduces \ac{FA} and \ac{DFA} accuracy, particularly in deeper networks, whereas \ac{BP} is less affected. Increasing the retained fraction recovers much of the benefit of full-rank Muon. Thus, the benefit of retaining more directions cannot be attributed simply to a larger norm of the update.

A complementary control tests whether adding arbitrary directions is sufficient. We added isotropic Gaussian noise to the parameter updates, with noise norm equal to 0.1 or 1 times the base update norm, which artificially increases the effective rank of the update. Adding noise does not reproduce Muon's accuracy gains in \ac{FA} or \ac{DFA} (\figref{fig:fig_muon} F). Together with the norm-matched rank control, it supports the interpretation that distributing updates across directions present in the learning signal is useful, while arbitrary high-dimensional motion alone is insufficient.

\begin{wrapfigure}{r}{0.5\columnwidth}
  \centering
  \includegraphics[width=0.5\columnwidth]{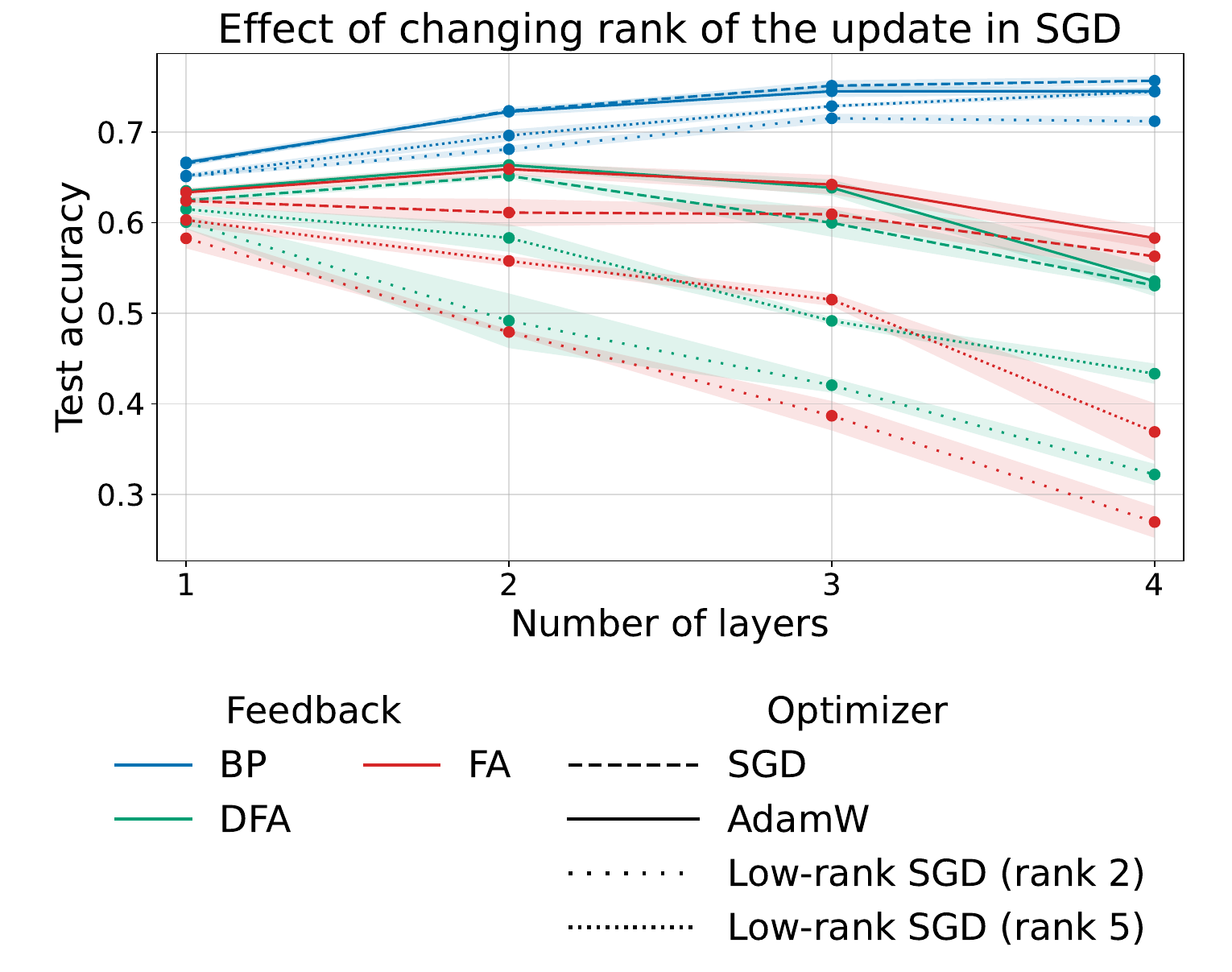}
  \caption{\textbf{Using the top-K directions as the opposite of Muon.} Using low-rank SGD results in much worse performance with \ac{FA}, whereas the effect is not so large with \ac{BP}.}
  \label{fig:top_k_sgd}
\end{wrapfigure}

Given that an optimiser that increases the dimensionality of the update improves the performance of \ac{FA} models. We investigated how an optimiser that does the opposite affects \ac{FA}. To do so, we replaced the orthogonalisation function in Muon with the rank r approximation of the momentum matrix using the SVD of the momentum. We used the rank 2 and rank 5 approximation (\figref{fig:top_k_sgd}) and found that while \ac{BP} models are not so affected, the effect on \ac{FA} models is catastrophic.

Taken together, these results support the role of update geometry in improving random-feedback learning. Muon allows weaker singular directions to contribute to learning, and the rank control shows that removing these directions can reintroduce the degradation with depth even when the update norm is held fixed.

\paragraph{Normalising the activities}

A second way to increase the dimensionality of the updates is to act directly on the activities of the network. 
As the gradient is the outer product of the error signal and the previous activation, increasing the dimensionality of the activation may increase the dimensionality of the gradients. 
\ac{BN} \citep{ioffe2015batch}, has been shown to promote more orthogonal hidden representations in deep networks \citep{daneshmand2020batch,daneshmand2021batch}, suggesting that it may prevent the collapse of the representations onto low-dimensional subspaces. 
Since our previous results indicate that \ac{FA} fails partly because its gradients and features become too low-dimensional, we added \ac{BN} layers after each convolutional layer, using the same experimental setup as above. 

\begin{figure}
    \centering
    \subfloat{\includegraphics[width=0.95\linewidth]{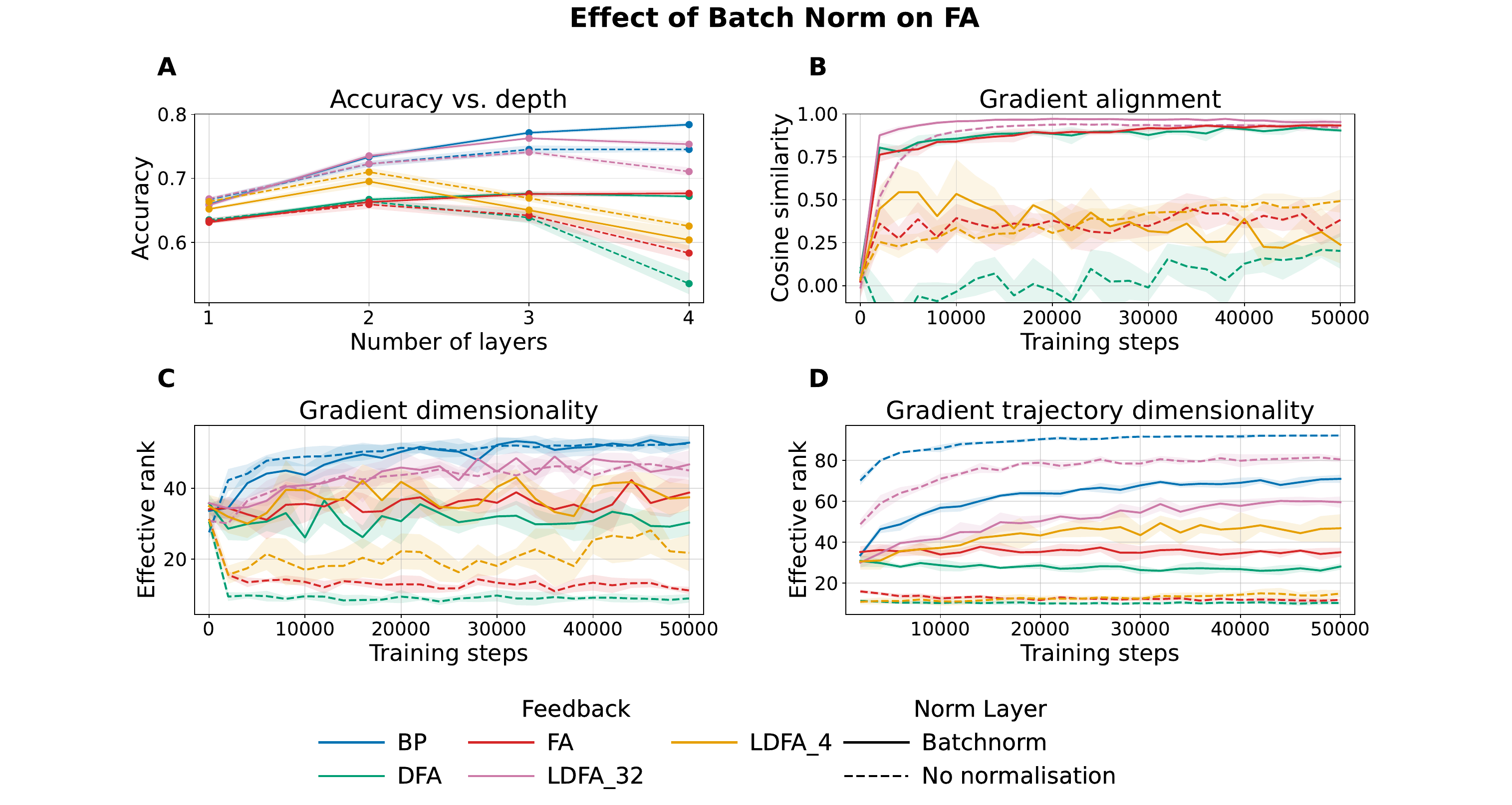}}
    \caption{\textbf{Optimising \ac{FA} networks with \ac{BN} results in better \ac{FA}.} \textbf{A} Effect of adding batch normalisation layers on the accuracy of \ac{FA} networks. The accuracy of models trained without \ac{BN} decreases, while the accuracy of models trained with \ac{BN} does not. \textbf{B-D} Metrics for the last layer of the 4 convolutional layer model. \textbf{B} Models with \ac{BN} layers have higher and smoother gradient alignment. \textbf{C} The gradients of models with \ac{BN} have higher dimensionality. \textbf{D} The trajectory of the gradients is also higher for models with \ac{BN}.}
    \label{fig:fig_batch_norm}
\end{figure}

The results were very similar to those obtained with Muon (\figref{fig:fig_batch_norm}). \ac{BN} substantially reduced the degradation in accuracy with depth (\figref{fig:fig_batch_norm} A), and increased gradient alignment (\figref{fig:fig_batch_norm} B).
Consistent with our hypothesis, it also increased the dimensionality of the gradients (\figref{fig:fig_batch_norm} C), the gradient trajectory  (\figref{fig:fig_batch_norm} D). 
Thus, improving the geometry of the representations can mitigate the main pathologies of \ac{FA}, further supporting the view that the poor performance of \ac{FA} in deep \acp{CNN} is due to low-dimensional updates not allowing the alignment process to continue.

Together, these results may explain why \ac{BN} and Muon produce similar improvements in \ac{FA}: both help maintain a richer set of directions through which alignment and learning can continue. 
Conversely, forcing updates to be low-rank severely harms FA, confirming that higher-dimensional optimisation is crucial for robust \ac{FA} training.

\section{Scaling to deeper models}
\label{sec:scaling}

\begin{table}[!htbp]
\caption{\textbf{Scaling \ac{FA} and \ac{DFA} to deeper architectures and harder datasets.}
Top-1 accuracy (\%, mean $\pm$ standard deviation across seeds) of AlexNet and ResNet-18 on STL-10, CIFAR-100, and ImageNet. For each feedback method, we compare AdamW \citep{loshchilov2017decoupled}, \ac{BN}, Muon, and Muon+\ac{BN}. Bold indicates the best mean accuracy within each feedback method for each dataset and architecture. NA indicates that the loss diverged for all seeds during the first epoch.}
\label{tab:scale}
\centering
\small
\setlength{\tabcolsep}{3pt}
\resizebox{\columnwidth}{!}{%

\begin{tabular}{llrr|rr|rr}
\toprule
 & Dataset & \multicolumn{2}{c}{STL-10} & \multicolumn{2}{c}{CIFAR-100} & \multicolumn{2}{c}{ImageNet} \\
 & Network & AlexNet & ResNet-18 & AlexNet & ResNet-18 & AlexNet & ResNet-18 \\
Feedback & Setting &  &  &  &  &  &  \\
\midrule
\multirow[t]{4}{*}{BP} & Baseline & 71.1 $\pm$ 0.7 & 72.2 $\pm$ 1.5 & 58.6 $\pm$ 0.3 & 63.1 $\pm$ 1.1 & 53.0 $\pm$ 0.1 & 60.3 $\pm$ 0.2 \\
 & BN & 80.8 $\pm$ 0.3 & 78.2 $\pm$ 0.6 & 68.3 $\pm$ 0.3 & 73.7 $\pm$ 0.4 & 53.1 $\pm$ 0.2 & 67.1 $\pm$ 0.1 \\
 & Muon & 85.2 $\pm$ 0.2 & 81.6 $\pm$ 0.4 & 66.0 $\pm$ 0.5 & 69.9 $\pm$ 0.4 & 53.2 $\pm$ 0.1 & 65.9 $\pm$ 0.2 \\
 & Muon + BN & \textbf{86.1 $\pm$ 0.3} & \textbf{84.1 $\pm$ 0.1} & \textbf{69.3 $\pm$ 0.3} & \textbf{75.3 $\pm$ 0.3} & \textbf{56.6 $\pm$ 0.1} & \textbf{68.2 $\pm$ 0.2} \\
\cline{1-8}
\multirow[t]{4}{*}{FA} & Baseline & 49.0 $\pm$ 1.9 & 10.2 $\pm$ 0.4 & 13.2 $\pm$ 1.2 & 1.1 $\pm$ 0.1 & 0.1 $\pm$ 0.0 & NA \\
 & BN & 69.7 $\pm$ 0.9 & 54.2 $\pm$ 0.8 & 45.0 $\pm$ 0.3 & 34.5 $\pm$ 0.8 & 11.3 $\pm$ 0.1 & 19.0 $\pm$ 0.1 \\
 & Muon & 70.8 $\pm$ 0.5 & 59.8 $\pm$ 1.1 & 45.9 $\pm$ 0.5 & 34.7 $\pm$ 0.7 & 11.7 $\pm$ 0.1 & 6.6 $\pm$ 0.2 \\
 & Muon + BN & \textbf{76.4 $\pm$ 0.6} & \textbf{63.0 $\pm$ 0.5} & \textbf{49.0 $\pm$ 0.5} & \textbf{46.8 $\pm$ 0.4} & \textbf{20.2 $\pm$ 0.2} & \textbf{27.4 $\pm$ 0.2} \\
\cline{1-8}
\multirow[t]{4}{*}{DFA} & Baseline & 47.3 $\pm$ 1.1 & 11.0 $\pm$ 0.8 & 21.4 $\pm$ 1.5 & 1.2 $\pm$ 0.2 & 0.2 $\pm$ 0.0 & NA \\
 & BN & 69.1 $\pm$ 0.9 & 67.3 $\pm$ 0.3 & 47.7 $\pm$ 0.4 & 36.9 $\pm$ 0.3 & 16.1 $\pm$ 0.1 & 14.9 $\pm$ 0.2 \\
 & Muon & 66.6 $\pm$ 1.2 & 60.0 $\pm$ 0.8 & 46.4 $\pm$ 0.3 & 10.3 $\pm$ 0.2 & 18.9 $\pm$ 0.1 & 2.1 $\pm$ 0.2 \\
 & Muon + BN & \textbf{74.3 $\pm$ 0.4} & \textbf{70.9 $\pm$ 0.6} & \textbf{50.3 $\pm$ 0.6} & \textbf{41.3 $\pm$ 0.5} & \textbf{26.2 $\pm$ 0.1} & \textbf{17.6 $\pm$ 0.2} \\
\bottomrule
\end{tabular}

}
\end{table}

Given that Muon and \ac{BN} improved the trainability of \ac{FA} networks in our previous setting, we next asked whether these interventions also scaled to harder datasets and more realistic architectures. We trained AlexNet \citep{krizhevsky2012imagenet} and ResNet-18 \citep{he2016deep} on STL-10 \citep{coates2011analysis}, CIFAR-100 \citep{krizhevsky2009learning}, and ImageNet \citep{russakovsky2015imagenet}. For each dataset and architecture, we compared \ac{BP}, \ac{FA}, and \ac{DFA} under four settings: AdamW \citep{loshchilov2017decoupled}, \ac{BN}, Muon, and Muon combined with \ac{BN} (Table~\ref{tab:scale}).

Consistent with the results from the previous section, baseline \ac{FA} performed poorly in deeper models. The degradation was especially severe for ResNet-18, where baseline \ac{FA} reached only 10.2\% on STL-10 and 1.1\% on CIFAR-100. Both \ac{BN} and Muon substantially improved \ac{FA} performance in all reported dataset architecture combinations. On STL-10, AlexNet trained with baseline \ac{FA} reached 49.0\%, whereas adding \ac{BN} increased performance to 69.7\%, Muon to 70.8\%, and Muon+\ac{BN} to 76.4\%. For ResNet-18, Muon+\ac{BN} improved performance from 10.2\% to 63.0\%.

The same trend held on CIFAR-100. Baseline \ac{FA} reached 13.2\% with AlexNet and 1.1\% with ResNet-18. Adding either \ac{BN} or Muon produced gains, and their combination gave the best performance, reaching 49.0\% and 46.8\%. \Ac{DFA} also benefited from both interventions: Muon+\ac{BN} increased STL-10 accuracy from 47.3\% to 74.3\% for AlexNet and from 11.0\% to 70.9\% for ResNet-18. On CIFAR-100, the corresponding improvements were from 21.4\% to 50.3\% and from 1.2\% to 41.3\%. The relative benefit of each intervention varied. For ResNet-18 trained with \ac{DFA} on CIFAR-100, \ac{BN} alone reached 36.9\%, compared with 10.3\% for Muon alone.

On ImageNet, baseline AlexNet performance was only 0.1\% with \ac{FA} and 0.2\% with \ac{DFA}. With \ac{FA}, \ac{BN} increased accuracy to 11.3\%, Muon to 11.7\%, and Muon+\ac{BN} to 20.2\%. With \ac{DFA}, the corresponding accuracies were 16.1\%, 18.9\%, and 26.2\%. Thus, the combined intervention enabled both methods to learn on ImageNet, although a substantial gap remained to AlexNet trained with \ac{BP}, which reached 56.6\% with Muon+\ac{BN}. For the ResNet-18, baseline \ac{FA} and \ac{DFA} diverged on ImageNet. Muon alone did not perform as well on ImageNet, reaching 6.6\%. Still, Muon combined with \ac{BN} performed best, reaching 27.4\% with \ac{FA} and 17.6\% with \ac{DFA}, respectively.

Muon and \ac{BN} were complementary for both \ac{FA} and \ac{DFA}: their combination gave the highest mean accuracy for each benchmark combination. Muon directly modifies the geometry of the parameter update by flattening the spectrum of the momentum, whereas \ac{BN} acts on the forward representations and changes the geometry of the gradients. The accuracy results are consistent with the benefits of preserving richer learning directions observed in the previous section.

\section{Discussion}

While feedback alignment was surprisingly shown to work in shallow MLP \citep{lillicrap2016random}, the fact that it did not scale well to deeper architectures raised doubts on whether the brain could be functioning on this principle. Indeed, convolutional networks are not biologically plausible due to neurons sharing exact weights, even though they are inspired by the visual system \citep{bartunov2018assessing,pogodin2021towards}, this weight sharing is what makes optimisation easier with backpropagation. Therefore, it would be desirable that an approximation to \ac{BP} also performs well with such architectures. In this work, we presented ways of making \ac{FA} work in deeper \acp{CNN}. Using a different optimiser or by normalising the activities, we found that \ac{FA} did not have such a steep decrease in accuracy when adding layers as it does when using standard methods. While we did not investigate why alignment happens, as previous work already investigated this question \citep{lillicrap2016random,refinetti2021align}, we focused on why it seemingly stops after a period of early alignment, and explained how to allow alignment to continue after this period.

Biological plausibility directly motivates the study of \ac{FA}, but the interventions used here should not be interpreted as biologically plausible learning algorithms in their current form. In particular, Muon requires orthogonalising a matrix-valued momentum update, which depends on coordinated information about the spectrum of an entire weight-update matrix rather than on purely synapse-local quantities. Although local plasticity rules such as Oja-like mechanisms can extract principal directions \citep{oja1982simplified,sanger1989optimal}, our implementation of Muon should be viewed as a probe of an alternative rule for updates, not as a direct model of synaptic plasticity.

The other mechanism we investigated was normalising the activities, in our case by using \ac{BN} \citep{ioffe2015batch}. Beyond its original motivation as a way of reducing internal covariate shift, \ac{BN} can be understood as helping deep networks preserve high-dimensional representations across layers. In unnormalised deep networks, the activity matrix over a batch can progressively collapse to a low-rank representation \citep{huh2021low,daneshmand2020batch}. By contrast, \ac{BN} counteracts this rank collapse and helps maintain multiple active directions of variation in the hidden activities \citep{daneshmand2020batch,daneshmand2021batch}. In addition, \ac{BN} often makes training less sensitive to initialisation, permits the use of larger learning rates, and improves gradient propagation \citep{ioffe2015batch,santurkar2018does}. While we did not explicitly control for these factors, both Muon and \ac{BN} improved \ac{FA} and \ac{DFA} in all reported settings in Table~\ref{tab:scale}, and their combination gave the highest mean accuracy. Their individual benefits varied across settings, so these results are consistent with complementary effects on learning rather than establishing a single shared mechanism. Although standard \ac{BN} is not strictly biologically plausible because it relies on batch-level statistics and global normalisation across examples, normalisation mechanisms are widespread in biological neural systems \citep{carandini2012normalization}. In particular, divisive normalisation and homeostatic regulation \citep{wen2024keeping,shen2021correspondence} could provide more local approximations that similarly stabilise activity scales and prevent representational collapse. Transformations of neural activity at the population level have been studied in the visual cortex \citep{failor2025visual}, and are thought to be behind its high dimensionality \citep{rigotti2013importance,fusi2016neurons,stringer2019high,failor2025visual}.

This perspective is also consistent with a growing body of work studying the dimensionality of neural population activity in the visual cortex. Rather than being confined to a small number of easily interpretable axes, cortical activity often occupies high-dimensional, structured subspaces, which can support flexible downstream readout and improve the separability of behaviourally relevant variables \citep{rigotti2013importance,fusi2016neurons,failor2025visual}. This has been studied particularly clearly in the visual cortex, where large-scale recordings in awake mice showed that responses to natural images span many dimensions and exhibit a scale-free eigenspectrum, suggesting that visual representations preserve a rich set of stimulus-related directions rather than collapsing onto a few dominant modes \citep{stringer2019high}. This high dimensionality is what would allow \ac{FA} alignment to work well.

Beyond biological plausibility, feedback alignment may be useful as a distinct optimisation system from standard backpropagation. Because FA uses fixed random feedback, it follows different parameter-space trajectories and can expose how optimisation techniques behave when the learning signal is only approximate. This makes FA a useful testbed for studying whether methods such as normalisation or update orthogonalisation improve learning by better conditioning BP gradients specifically, or by more generally shaping update geometry. Moreover, alternative credit-assignment rules can affect loss-landscape exploration \citep{pascanu2025optimizers}, for example, \ac{FA} could use structured feedback that guides learning to different solutions than what \ac{BP} would reach. For instance, \ac{FA} has been shown to allow better continual learning \citep{folchini2025exploring}.

\section*{Acknowledgements}
This work was supported by Wellcome Trust 200790/Z/16/Z, Simons Foundation 564408 and EPSRC EP/R035806/1.

\bibliographystyle{plainnat}
\bibliography{references}

@inproceedings{boopathy2022train,
  title={How to train your wide neural network without backprop: An input-weight alignment perspective},
  author={Boopathy, Akhilan and Fiete, Ila},
  booktitle={International Conference on Machine Learning},
  pages={2178--2205},
  year={2022},
  organization={PMLR}
}

@misc{jordan2024muon,
  author       = {Keller Jordan and Yuchen Jin and Vlado Boza and Jiacheng You and
                  Franz Cesista and Laker Newhouse and Jeremy Bernstein},
  title        = {Muon: An optimizer for hidden layers in neural networks},
  year         = {2024},
  url          = {https://kellerjordan.github.io/posts/muon/}
}

@article{bernstein2024old,
  title={Old optimizer, new norm: An anthology},
  author={Bernstein, Jeremy and Newhouse, Laker},
  journal={arXiv preprint arXiv:2409.20325},
  year={2024}
}

@misc{bernstein2025deriving,
  author = {Jeremy Bernstein},
  title = {Deriving Muon},
  url = {https://jeremybernste.in/writing/deriving-muon},
  year = {2025}
}

@article{ba2016layer,
  title={Layer normalization},
  author={Ba, Jimmy Lei and Kiros, Jamie Ryan and Hinton, Geoffrey E},
  journal={arXiv preprint arXiv:1607.06450},
  year={2016}
}

@article{simonyan2014very,
  title={Very deep convolutional networks for large-scale image recognition},
  author={Simonyan, Karen and Zisserman, Andrew},
  journal={arXiv preprint arXiv:1409.1556},
  year={2014}
}

@inproceedings{refinetti2021align,
  title={Align, then memorise: the dynamics of learning with feedback alignment},
  author={Refinetti, Maria and d’Ascoli, St{\'e}phane and Ohana, Ruben and Goldt, Sebastian},
  booktitle={International Conference on Machine Learning},
  pages={8925--8935},
  year={2021},
  organization={PMLR}
}

@article{bartunov2018assessing,
  title={Assessing the scalability of biologically-motivated deep learning algorithms and architectures},
  author={Bartunov, Sergey and Santoro, Adam and Richards, Blake and Marris, Luke and Hinton, Geoffrey E and Lillicrap, Timothy},
  journal={Advances in neural information processing systems},
  volume={31},
  year={2018}
}

@article{launay2019principled,
  title={Principled training of neural networks with direct feedback alignment},
  author={Launay, Julien and Poli, Iacopo and Krzakala, Florent},
  journal={arXiv preprint arXiv:1906.04554},
  year={2019}
}

@inproceedings{he2016deep,
  title={Deep residual learning for image recognition},
  author={He, Kaiming and Zhang, Xiangyu and Ren, Shaoqing and Sun, Jian},
  booktitle={Proceedings of the IEEE conference on computer vision and pattern recognition},
  pages={770--778},
  year={2016}
}

@article{akrout2019deep,
  title={Deep learning without weight transport},
  author={Akrout, Mohamed and Wilson, Collin and Humphreys, Peter and Lillicrap, Timothy and Tweed, Douglas B},
  journal={Advances in neural information processing systems},
  volume={32},
  year={2019}
}

@article{xiao2018biologically,
  title={Biologically-plausible learning algorithms can scale to large datasets},
  author={Xiao, Will and Chen, Honglin and Liao, Qianli and Poggio, Tomaso},
  journal={arXiv preprint arXiv:1811.03567},
  year={2018}
}

@article{bardes2021vicreg,
  title={Vicreg: Variance-invariance-covariance regularization for self-supervised learning},
  author={Bardes, Adrien and Ponce, Jean and LeCun, Yann},
  journal={arXiv preprint arXiv:2105.04906},
  year={2021}
}

@inproceedings{ioffe2015batch,
  title={Batch normalization: Accelerating deep network training by reducing internal covariate shift},
  author={Ioffe, Sergey and Szegedy, Christian},
  booktitle={International conference on machine learning},
  pages={448--456},
  year={2015},
  organization={pmlr}
}

@misc{krizhevsky2009learning,
  title        = {Learning Multiple Layers of Features from Tiny Images},
  author       = {Krizhevsky, Alex},
  year         = {2009},
  howpublished = {Technical report, University of Toronto}
}

@article{crick1989recent,
  title={The recent excitement about neural networks},
  author={Crick, Francis},
  journal={Nature},
  volume={337},
  number={6203},
  pages={129--132},
  year={1989},
  publisher={Nature Publishing Group UK London}
}

@article{lillicrap2016random,
  title={Random synaptic feedback weights support error backpropagation for deep learning},
  author={Lillicrap, Timothy P and Cownden, Daniel and Tweed, Douglas B and Akerman, Colin J},
  journal={Nature communications},
  volume={7},
  number={1},
  pages={13276},
  year={2016},
  publisher={Nature Publishing Group UK London}
}

@article{rumelhart1986learning,
  title={Learning representations by back-propagating errors},
  author={Rumelhart, David E and Hinton, Geoffrey E and Williams, Ronald J},
  journal={nature},
  volume={323},
  number={6088},
  pages={533--536},
  year={1986},
  publisher={Nature Publishing Group UK London}
}

@article{lillicrap2020backpropagation,
  title={Backpropagation and the brain},
  author={Lillicrap, Timothy P and Santoro, Adam and Marris, Luke and Akerman, Colin J and Hinton, Geoffrey},
  journal={Nature Reviews Neuroscience},
  volume={21},
  number={6},
  pages={335--346},
  year={2020},
  publisher={Nature Publishing Group UK London}
}

@article{nokland2016direct,
  title={Direct feedback alignment provides learning in deep neural networks},
  author={N{\o}kland, Arild},
  journal={Advances in neural information processing systems},
  volume={29},
  year={2016}
}

@article{frenkel2021learning,
  title={Learning without feedback: Fixed random learning signals allow for feedforward training of deep neural networks},
  author={Frenkel, Charlotte and Lefebvre, Martin and Bol, David},
  journal={Frontiers in neuroscience},
  volume={15},
  pages={629892},
  year={2021},
  publisher={Frontiers Media SA}
}

@article{moskovitz2018feedback,
  title={Feedback alignment in deep convolutional networks},
  author={Moskovitz, Theodore H and Litwin-Kumar, Ashok and Abbott, LF},
  journal={arXiv preprint arXiv:1812.06488},
  year={2018}
}

@inproceedings{liao2016important,
  title={How important is weight symmetry in backpropagation?},
  author={Liao, Qianli and Leibo, Joel and Poggio, Tomaso},
  booktitle={Proceedings of the AAAI Conference on Artificial Intelligence},
  volume={30},
  year={2016}
}

@inproceedings{kunin2020two,
  title={Two routes to scalable credit assignment without weight symmetry},
  author={Kunin, Daniel and Nayebi, Aran and Sagastuy-Brena, Javier and Ganguli, Surya and Bloom, Jonathan and Yamins, Daniel},
  booktitle={International Conference on Machine Learning},
  pages={5511--5521},
  year={2020},
  organization={PMLR}
}

@article{cheon2024pretraining,
  title={Pretraining with random noise for fast and robust learning without weight transport},
  author={Cheon, Jeonghwan and Lee, Sang Wan and Paik, Se-Bum},
  journal={Advances in Neural Information Processing Systems},
  volume={37},
  pages={13748--13768},
  year={2024}
}

@article{liu2025muon,
  title={Muon is scalable for llm training},
  author={Liu, Jingyuan and Su, Jianlin and Yao, Xingcheng and Jiang, Zhejun and Lai, Guokun and Du, Yulun and Qin, Yidao and Xu, Weixin and Lu, Enzhe and Yan, Junjie and others},
  journal={arXiv preprint arXiv:2502.16982},
  year={2025}
}

@inproceedings{roy2007effective,
  title={The effective rank: A measure of effective dimensionality},
  author={Roy, Olivier and Vetterli, Martin},
  booktitle={2007 15th European signal processing conference},
  pages={606--610},
  year={2007},
  organization={IEEE}
}

@article{russakovsky2015imagenet,
  title={Imagenet large scale visual recognition challenge},
  author={Russakovsky, Olga and Deng, Jia and Su, Hao and Krause, Jonathan and Satheesh, Sanjeev and Ma, Sean and Huang, Zhiheng and Karpathy, Andrej and Khosla, Aditya and Bernstein, Michael and others},
  journal={International journal of computer vision},
  volume={115},
  number={3},
  pages={211--252},
  year={2015},
  publisher={Springer}
}

@article{huh2021low,
  title={The low-rank simplicity bias in deep networks},
  author={Huh, Minyoung and Mobahi, Hossein and Zhang, Richard and Cheung, Brian and Agrawal, Pulkit and Isola, Phillip},
  journal={arXiv preprint arXiv:2103.10427},
  year={2021}
}

@article{andriushchenko2023sharpness,
  title={Sharpness-aware minimization leads to low-rank features},
  author={Andriushchenko, Maksym and Bahri, Dara and Mobahi, Hossein and Flammarion, Nicolas},
  journal={Advances in Neural Information Processing Systems},
  volume={36},
  pages={47032--47051},
  year={2023}
}

@article{valle2018deep,
  title={Deep learning generalizes because the parameter-function map is biased towards simple functions},
  author={Valle-Perez, Guillermo and Camargo, Chico Q and Louis, Ard A},
  journal={arXiv preprint arXiv:1805.08522},
  year={2018}
}

@article{cogswell2015reducing,
  title={Reducing overfitting in deep networks by decorrelating representations},
  author={Cogswell, Michael and Ahmed, Faruk and Girshick, Ross and Zitnick, Larry and Batra, Dhruv},
  journal={arXiv preprint arXiv:1511.06068},
  year={2015}
}

@article{hanut2025training,
  title={Training Large Neural Networks With Low-Dimensional Error Feedback},
  author={Hanut, Maher and Kadmon, Jonathan},
  journal={arXiv preprint arXiv:2502.20580},
  year={2025}
}

@article{loshchilov2017decoupled,
  title={Decoupled weight decay regularization},
  author={Loshchilov, Ilya and Hutter, Frank},
  journal={arXiv preprint arXiv:1711.05101},
  year={2017}
}

@article{pogodin2021towards,
  title={Towards biologically plausible convolutional networks},
  author={Pogodin, Roman and Mehta, Yash and Lillicrap, Timothy and Latham, Peter E},
  journal={Advances in neural information processing systems},
  volume={34},
  pages={13924--13936},
  year={2021}
}

@article{oja1982simplified,
  title={Simplified neuron model as a principal component analyzer},
  author={Oja, Erkki},
  journal={Journal of mathematical biology},
  volume={15},
  number={3},
  pages={267--273},
  year={1982},
  publisher={Springer}
}

@article{sanger1989optimal,
  title={Optimal unsupervised learning in a single-layer linear feedforward neural network},
  author={Sanger, Terence D},
  journal={Neural networks},
  volume={2},
  number={6},
  pages={459--473},
  year={1989},
  publisher={Elsevier}
}

@article{daneshmand2020batch,
  title={Batch normalization provably avoids ranks collapse for randomly initialised deep networks},
  author={Daneshmand, Hadi and Kohler, Jonas and Bach, Francis and Hofmann, Thomas and Lucchi, Aurelien},
  journal={Advances in Neural Information Processing Systems},
  volume={33},
  pages={18387--18398},
  year={2020}
}

@article{daneshmand2021batch,
  title={Batch normalization orthogonalizes representations in deep random networks},
  author={Daneshmand, Hadi and Joudaki, Amir and Bach, Francis},
  journal={Advances in Neural Information Processing Systems},
  volume={34},
  pages={4896--4906},
  year={2021}
}

@article{santurkar2018does,
  title={How does batch normalization help optimization?},
  author={Santurkar, Shibani and Tsipras, Dimitris and Ilyas, Andrew and Madry, Aleksander},
  journal={Advances in neural information processing systems},
  volume={31},
  year={2018}
}

@article{krizhevsky2012imagenet,
  title={Imagenet classification with deep convolutional neural networks},
  author={Krizhevsky, Alex and Sutskever, Ilya and Hinton, Geoffrey E},
  journal={Advances in neural information processing systems},
  volume={25},
  year={2012}
}

@inproceedings{coates2011analysis,
  title={An analysis of single-layer networks in unsupervised feature learning},
  author={Coates, Adam and Ng, Andrew and Lee, Honglak},
  booktitle={Proceedings of the fourteenth international conference on artificial intelligence and statistics},
  pages={215--223},
  year={2011},
  organization={JMLR Workshop and Conference Proceedings}
}

@article{paszke2019pytorch,
  title={Pytorch: An imperative style, high-performance deep learning library},
  author={Paszke, Adam and Gross, Sam and Massa, Francisco and Lerer, Adam and Bradbury, James and Chanan, Gregory and Killeen, Trevor and Lin, Zeming and Gimelshein, Natalia and Antiga, Luca and others},
  journal={Advances in neural information processing systems},
  volume={32},
  year={2019}
}

@Misc{Yadan2019Hydra,
  author =       {Omry Yadan},
  title =        {Hydra - A framework for elegantly configuring complex applications},
  howpublished = {Github},
  year =         {2019},
  url =          {https://github.com/facebookresearch/hydra}
}

@Article{Hunter:2007,
  Author    = {Hunter, J. D.},
  Title     = {Matplotlib: A 2D graphics environment},
  Journal   = {Computing in Science \& Engineering},
  Volume    = {9},
  Number    = {3},
  Pages     = {90--95},
  publisher = {IEEE COMPUTER SOC},
  doi       = {10.1109/MCSE.2007.55},
  year      = 2007
}

@article{carandini2012normalization,
  title={Normalization as a canonical neural computation},
  author={Carandini, Matteo and Heeger, David J},
  journal={Nature reviews neuroscience},
  volume={13},
  number={1},
  pages={51--62},
  year={2012},
  publisher={Nature Publishing Group UK London}
}

@article{wen2024keeping,
  title={Keeping your brain in balance: homeostatic regulation of network function},
  author={Wen, Wei and Turrigiano, Gina G},
  journal={Annual review of neuroscience},
  volume={47},
  year={2024},
  publisher={Annual Reviews}
}

@article{rigotti2013importance,
  title={The importance of mixed selectivity in complex cognitive tasks},
  author={Rigotti, Mattia and Barak, Omri and Warden, Melissa R and Wang, Xiao-Jing and Daw, Nathaniel D and Miller, Earl K and Fusi, Stefano},
  journal={Nature},
  volume={497},
  number={7451},
  pages={585--590},
  year={2013},
  publisher={Nature Publishing Group UK London}
}

@article{fusi2016neurons,
  title={Why neurons mix: high dimensionality for higher cognition},
  author={Fusi, Stefano and Miller, Earl K and Rigotti, Mattia},
  journal={Current opinion in neurobiology},
  volume={37},
  pages={66--74},
  year={2016},
  publisher={Elsevier}
}

@article{failor2025visual,
  title={Visual experience orthogonalizes visual cortical stimulus responses via population code transformation},
  author={Failor, Samuel W and Carandini, Matteo and Harris, Kenneth D},
  journal={Cell Reports},
  volume={44},
  number={2},
  year={2025},
  publisher={Elsevier}
}

@article{stringer2019high,
  title={High-dimensional geometry of population responses in visual cortex},
  author={Stringer, Carsen and Pachitariu, Marius and Steinmetz, Nicholas and Carandini, Matteo and Harris, Kenneth D},
  journal={Nature},
  volume={571},
  number={7765},
  pages={361--365},
  year={2019},
  publisher={Nature Publishing Group UK London}
}

@article{pascanu2025optimizers,
  title={Optimizers qualitatively alter solutions and we should leverage this},
  author={Pascanu, Razvan and Lyle, Clare and Modoranu, Ionut-Vlad and Borras, Naima Elosegui and Alistarh, Dan and Velickovic, Petar and Chandar, Sarath and De, Soham and Martens, James},
  journal={arXiv preprint arXiv:2507.12224},
  year={2025}
}

@article{folchini2025exploring,
  title={Exploring the potential of Direct Feedback Alignment for Continual Learning},
  author={Folchini, Sara and Arora, Viplove and Goldt, Sebastian},
  journal={Transactions on Machine Learning Research},
  year={2025}
}

@article{shen2021correspondence,
  title={A correspondence between normalization strategies in artificial and biological neural networks},
  author={Shen, Yang and Wang, Julia and Navlakha, Saket},
  journal={Neural computation},
  volume={33},
  number={12},
  pages={3179--3203},
  year={2021},
  publisher={MIT Press One Rogers Street, Cambridge, MA 02142-1209, USA journals-info~…}
}

@article{shumaylov2026muon,
  title={Muon is Not That Special: Random or Inverted Spectra Work Just as Well},
  author={Shumaylov, Zakhar and Da Costa, Natha{\"e}l and Zaika, Peter and Mucs{\'a}nyi, B{\'a}lint and Massucco, Alex and Gelberg, Yoav and Sch{\"o}nlieb, Carola-Bibiane and Gal, Yarin and Hennig, Philipp},
  journal={arXiv preprint arXiv:2605.11181},
  year={2026}
}

@article{girotti2021convergence,
  title={Convergence analysis and implicit regularization of feedback alignment for deep linear networks},
  author={Girotti, Manuela and Mitliagkas, Ioannis and Gidel, Gauthier},
  journal={arXiv preprint arXiv:2110.10815},
  year={2021}
}

@article{robertson2023implicit,
  title={Implicit regularization in feedback alignment learning mechanisms for neural networks},
  author={Robertson, Zachary and Koyejo, Oluwasanmi},
  journal={arXiv preprint arXiv:2306.01870},
  year={2023}
}

@article{saxe2013exact,
  title={Exact solutions to the nonlinear dynamics of learning in deep linear neural networks},
  author={Saxe, Andrew M and McClelland, James L and Ganguli, Surya},
  journal={arXiv preprint arXiv:1312.6120},
  year={2013}
}

@article{fukumizu1998effect,
  title={Effect of batch learning in multilayer neural networks},
  author={Fukumizu, Kenji},
  journal={Gen},
  volume={1},
  number={04},
  pages={1E--03},
  year={1998}
}

@article{arora2019implicit,
  title={Implicit regularization in deep matrix factorization},
  author={Arora, Sanjeev and Cohen, Nadav and Hu, Wei and Luo, Yuping},
  journal={Advances in neural information processing systems},
  volume={32},
  year={2019}
}

@inproceedings{arora2018optimization,
  title={On the optimization of deep networks: Implicit acceleration by overparameterization},
  author={Arora, Sanjeev and Cohen, Nadav and Hazan, Elad},
  booktitle={International conference on machine learning},
  pages={244--253},
  year={2018},
  organization={PMLR}
}

@article{ji2018gradient,
  title={Gradient descent aligns the layers of deep linear networks},
  author={Ji, Ziwei and Telgarsky, Matus},
  journal={arXiv preprint arXiv:1810.02032},
  year={2018}
}

@article{du2018algorithmic,
  title={Algorithmic regularization in learning deep homogeneous models: Layers are automatically balanced},
  author={Du, Simon S and Hu, Wei and Lee, Jason D},
  journal={Advances in neural information processing systems},
  volume={31},
  year={2018}
}

@misc{offergeld2026correlationsruininggradientdescent,
      title={Correlations Are Ruining Your Gradient Descent}, 
      author={Joshua Offergeld and Nasir Ahmad},
      year={2026},
      eprint={2407.10780},
      archivePrefix={arXiv},
      primaryClass={cs.LG},
      url={https://arxiv.org/abs/2407.10780}, 
}

@article{joudaki2023impact,
  title={On the impact of activation and normalization in obtaining isometric embeddings at initialization},
  author={Joudaki, Amir and Daneshmand, Hadi and Bach, Francis},
  journal={Advances in Neural Information Processing Systems},
  volume={36},
  pages={39855--39875},
  year={2023}
}

\newpage
\appendix
\section{Feedback methods and implementation}
\label{sec:feedback_implementation}
In this section, we describe the implementation details we omitted in the main text of the four feedback methods: \ac{BP}, \ac{FA}, \ac{DFA}, and \ac{LDFA}, presented in the main text and illustrated in \figref{fig:feedback_methods}. All four methods use the same forward operations and compute the task loss at the output. They differ in how the output error reaches the hidden layers and whether the feedback weights are tied to the forward weights, fixed, or learned.

\paragraph{Backpropagation.}
\Ac{BP} \citep{rumelhart1986learning} uses the transposed forward weights to propagate the output error back to the hidden layers. This is the standard method for training neural networks and usually serves as an upper bound for comparison with feedback methods.

\paragraph{Feedback alignment.}
\Ac{FA} \citep{lillicrap2016random} replaces the input derivative of each linear or convolutional operation by a projection through a fixed random map. Feedback tensors are initialised with Kaiming-uniform sampling, same as the forward weights. 

\paragraph{Direct feedback alignment.}
\Ac{DFA} \citep{nokland2016direct} bypasses the intervening feedback path. In our implementation, the feedback is sent after the block's normalisation, activation, and pooling. The projection is reshaped to the block output, and differentiation through those local operations then produces the convolutional parameter updates.

Following \citep{launay2019principled}, the direct maps share one fixed random matrix. If the output has $C$ classes and the largest hidden state has $m_{\max}$ elements, we sample $R\in\mathbb{R}^{C\times m_{\max}}$ with independent entries uniform on $[-m_{\max}^{-1/2},m_{\max}^{-1/2}]$. For a hidden state with $m_\ell$ elements, the effective map is
\begin{equation}
D_\ell=\sqrt{\frac{m_{\max}}{m_\ell}}\,\bigl(R_{:,1:m_\ell}\bigr)^\top.
\label{eq:dfa_shared_projection}
\end{equation}
Thus, the maps have the appropriate hidden-size scaling but share entries across layers, rather than being independently sampled. For a convolutional block, $m_\ell$ includes both channels and spatial positions after pooling. It is necessary to have such a shared structure as using distinct feedback maps uses too much memory. Moreover, as explained in \citep{refinetti2021align}, it is fine to have aligned projections, as such alignment naturally occurs in linear \ac{BP} networks \citep{ji2018gradient}. We also experimented with independent projections for each layer on CIFAR10, but found that this did not improve performance.

\paragraph{Low-dimensional feedback alignment.}
We followed the local formulation of LDFA  presented in \citet{hanut2025training}.  LDFA propagates errors between adjacent layers using a factorised map $B_\ell=Q_\ell P_\ell$. For $W_\ell\in\mathbb{R}^{n_\ell\times n_{\ell-1}}$, we use $Q_\ell\in\mathbb{R}^{n_{\ell-1}\times r_\ell}$ and $P_\ell\in\mathbb{R}^{r_\ell\times n_\ell}$, so the dense feedback map has rank at most $r_\ell$. 
Both factors are learned alongside the forward weights using local signals. An Oja-type rule adapts $P_\ell$ to the leading directions of the incoming error's second moment, while a Hebbian rule updates $Q_\ell$ using the correlation between $h_{\ell-1}$ and the compressed error $P_\ell\delta_\ell^{LDFA}$. This allows the feedback subspace to adapt during training while retaining the dimensionality constraint.
$Q_\ell$ is initialised using Kaiming-uniform and $P_\ell$ starts with orthonormal rows. We trained them using Adamw and included the learning rate  in the hyperparameter search. For convolutional layers, $P_\ell$ is a shared $1\times1$ projection from $C_{\rm out}$ error channels to $r_\ell$ channels, followed by the adjoint operation associated with a learned spatial kernel $Q_\ell$.

\section{Biologically plausible methods}

As mentioned in the discussion, the interventions used in the main text are not biologically plausible in their current implementation. We also tried two biologically plausible methods with low computational cost to increase the dimensionality of the updates: adding a local loss at each layer and using \ac{LN} \citep{ba2016layer}. The results appear similar to those obtained with Muon and \ac{BN}.

\subsection{Local loss}
\begin{figure}[!htbp]
    \centering
    \subfloat{\includegraphics[width=0.9\linewidth]{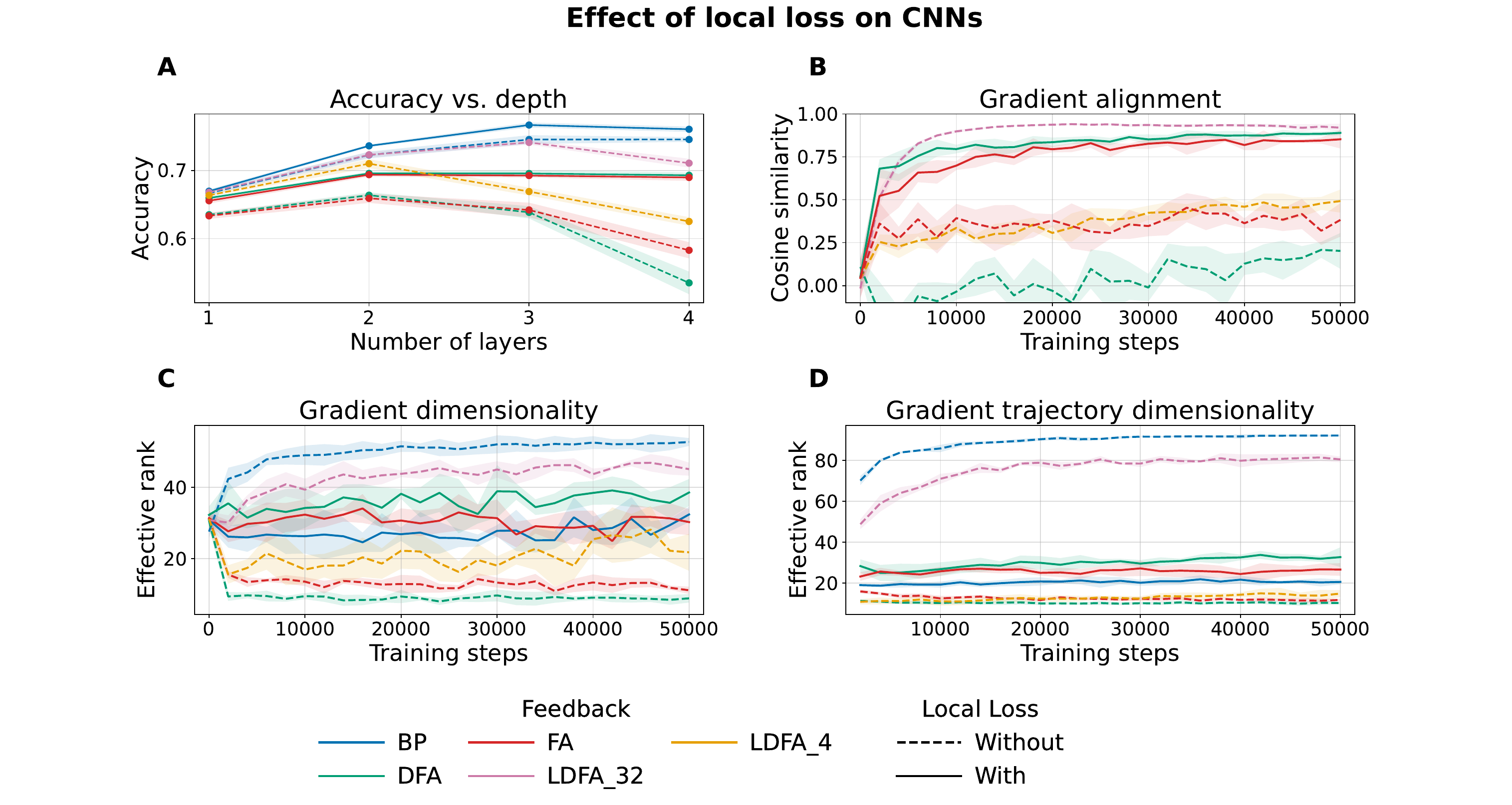}}
    \caption{\textbf{Adding a local loss to the \ac{FA} gradients.} \textbf{A} Adding layers on the accuracy of \ac{FA} networks. The accuracy of models trained with standard optimisers decreases, while those trained with the local loss do not. \textbf{B-D} Metrics for the 4 convolutional layer model. \textbf{B}  Models trained with local loss have smoother and higher gradient alignment. \textbf{C} The gradients of models trained with the local loss have higher effective rank. \textbf{D} The trajectory of the gradients is also higher for models trained with the local loss.}
    \label{fig:local_loss}
\end{figure}

We added a local loss at each layer, aimed at making the activity dimensionality higher. Specifically, we used a loss introduced in \citep{cogswell2015reducing} that minimises the covariance of the channels inside a layer; we did not backpropagate the loss of one layer to the previous layers. We followed the formulation used in \citep{bardes2021vicreg}: 
\begin{equation}
\mathcal{L}_{\mathrm{decor}}(X)
=
\frac{1}{D}
\sum_{\substack{i,j=1 \\ i \neq j}}^{D}
\left[
\frac{1}{B-1}
\sum_{b=1}^{B}
\left(x_{bi} - \bar{x}_i\right)
\left(x_{bj} - \bar{x}_j\right)
\right]^2,
\bar{x}_i = \frac{1}{B}\sum_{b=1}^{B} x_{bi}
\end{equation}
It minimises the off-diagonal components of the covariance of the activity in a given layer for one batch. The goal of this loss is to make the dimensionality higher by reducing the redundancies of the representations inside a layer. It is related to a lower bound approximation to the rank of a matrix introduced in \citep{daneshmand2020batch} $r(X)=\frac{\operatorname{Tr}(M(X))^2}{\|M(X)\|_F^2}, M(X) = \frac{X^\top X}{N}$, as minimising the off diagonal components of the covariance will have a similar objective to maximising the ratio of the squared trace to the squared frobenius norm of the covariance. We add the gradient of this loss weighted by $\lambda$ to the gradient of the cross-entropy loss used for classification. We optimised the network using AdamW \citep{loshchilov2017decoupled}.

We ran the same analysis as the one we did on Muon, and the results appear similar (\figref{fig:local_loss}). The accuracy does not decrease much when adding layers (\figref{fig:local_loss} A). When looking at the 4 convolutional layers network, we also see better behavior (\figref{fig:local_loss} B-D). The gradients are more aligned, and the alignment is more stable (\figref{fig:local_loss} B). The geometry of the updates is also better behaved. The gradients are higher-dimensional (\figref{fig:local_loss} C), and their trajectory is also higher-dimensional (\figref{fig:local_loss} D). 
The local loss did not scale well when using the deeper networks on the datasets we tested.

\subsection{Layer normalisation}

\begin{figure}[!htbp]
    \centering
    \subfloat{\includegraphics[width=0.9\linewidth]{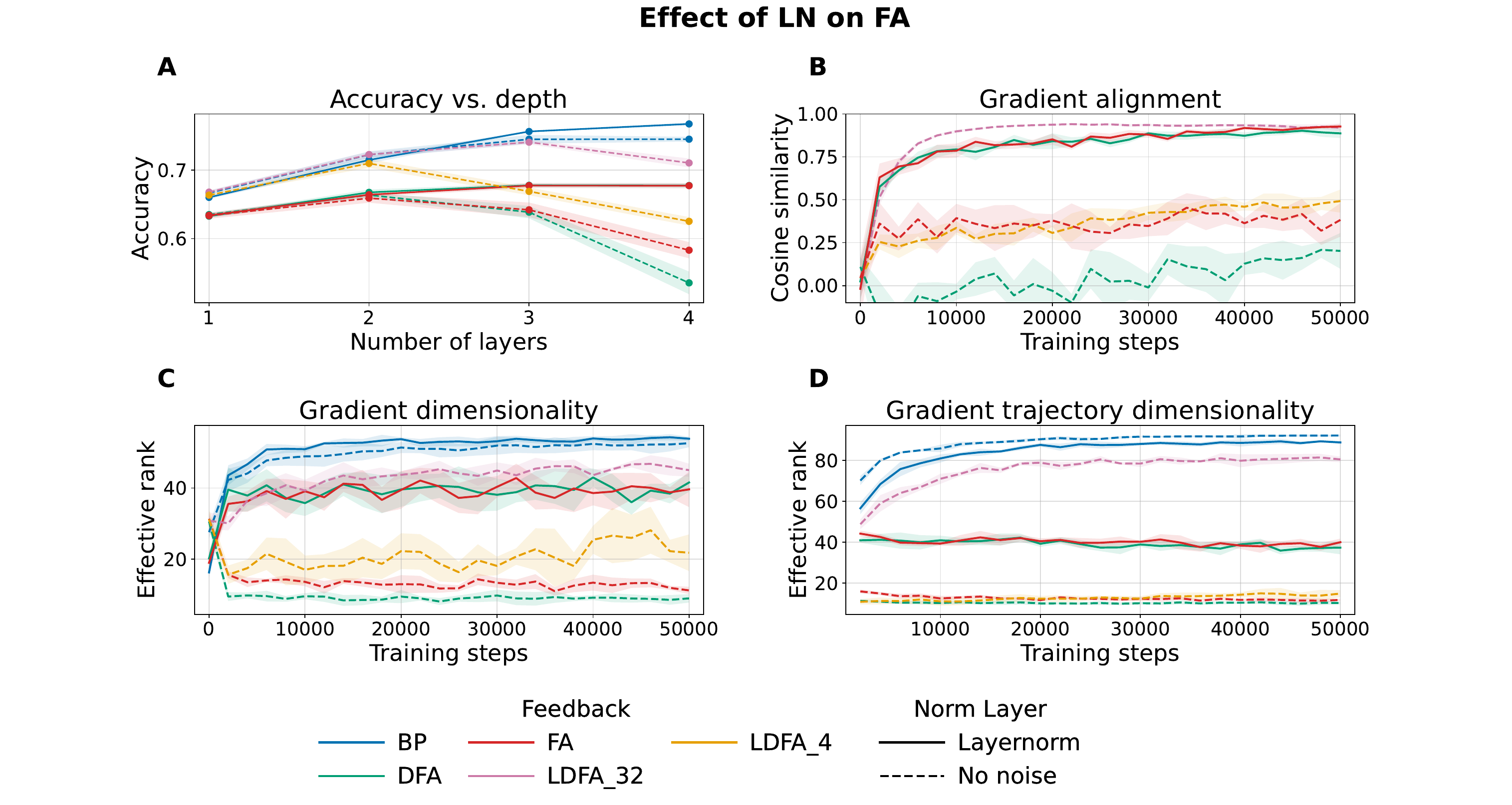}}
    \caption{\textbf{Layer normalisation improves the trainability of \ac{FA} networks.} \textbf{A} Accuracy as a function of network depth. Adding \ac{LN} reduces the loss of accuracy with increasing depth for \ac{FA} and \ac{DFA}. \textbf{B-D} Metrics for the last convolutional layer of the four convolutional-layer network. \textbf{B} With \ac{LN}, \ac{FA} and \ac{DFA} have higher and more stable gradient alignment. \textbf{C} Their gradients maintain a higher effective rank. \textbf{D} Their gradient trajectories also have higher effective rank.}
    \label{fig:layer_norm}
\end{figure}

We also normalised the activities within each layer using \ac{LN} \citep{ba2016layer}. For each example, \ac{LN} computes the mean and variance across the units in the layer, without using statistics from other examples in the mini-batch, which makes it more biologically plausible. Unlike the local loss, \ac{LN} does not explicitly minimise correlations between units, however, similarly to \ac{BN}, \ac{LN} was found to promote isometry in the activities \citep{joudaki2023impact}. 

We repeated the analysis used for the local loss and found similar effects for \ac{FA} and \ac{DFA} (\figref{fig:layer_norm}). Without normalisation, their accuracy decreases as the number of convolutional layers increases beyond two, whereas with \ac{LN} it remains approximately stable (\figref{fig:layer_norm} A). 
In the four convolutional-layer network, both methods achieve higher and more stable gradient alignment with \ac{LN} (\figref{fig:layer_norm} B). Their gradients also maintain a higher effective rank throughout training (\figref{fig:layer_norm} C), and their gradient trajectories span a higher-dimensional subspace (\figref{fig:layer_norm} D).
These results are consistent with the view that normalising the activities helps sustain alignment by preventing the learning signal from becoming confined to a small number of directions.
Table~\ref{tab:layer_norm} extends the comparison of \ac{LN} and Muon to AlexNet and ResNet-18 on STL-10, CIFAR-100, and Tiny ImageNet, a subset of ImageNet \citep{russakovsky2015imagenet}. \Ac{LN} performs slightly worse than \ac{BN} in general.  The combination of Muon and \ac{LN} still achieves the highest reported mean accuracy for each benchmark pair.

\begin{table}[!htbp]
\caption{\textbf{Layer normalisation and Muon in AlexNet and ResNet-18.}
Top-1 accuracy (\%, mean $\pm$ standard deviation across seeds) on STL-10, CIFAR-100, and Tiny ImageNet for  \ac{BP}, \ac{FA} and \ac{DFA}. Baseline uses AdamW \citep{loshchilov2017decoupled} without normalisation; LN adds layer normalisation, Muon changes the optimiser, and Muon + LN combines both interventions. Bold marks the highest mean accuracy within each feedback method for each dataset and architecture.}
\label{tab:layer_norm}
\centering
\small
\setlength{\tabcolsep}{3pt}
\resizebox{\columnwidth}{!}{
\begin{tabular}{llrr|rr|rr}
\toprule
 & Dataset & \multicolumn{2}{c}{STL-10} & \multicolumn{2}{c}{CIFAR-100} & \multicolumn{2}{c}{Tiny ImageNet} \\
 & Network & AlexNet & ResNet-18 & AlexNet & ResNet-18 & AlexNet & ResNet-18 \\
Feedback & Setting &  &  &  &  &  &  \\
\midrule
\multirow[t]{5}{*}{BP} & Baseline & 71.1 $\pm$ 0.7 & 72.2 $\pm$ 1.5 & 58.6 $\pm$ 0.3 & 63.1 $\pm$ 1.1 & 38.2 $\pm$ 0.4 & 31.0 $\pm$ 0.4 \\
 & LN & 74.6 $\pm$ 0.3 & 70.7 $\pm$ 0.4 & 65.6 $\pm$ 0.2 & 68.6 $\pm$ 0.4 & 47.8 $\pm$ 0.1 & 39.1 $\pm$ 0.3 \\
 & Muon & 85.2 $\pm$ 0.2 & 81.6 $\pm$ 0.4 & 66.0 $\pm$ 0.5 & 69.9 $\pm$ 0.4 & 46.0 $\pm$ 0.2 & 37.4 $\pm$ 0.1 \\
 & Muon + LN & \textbf{85.5 $\pm$ 0.3} & \textbf{83.0 $\pm$ 0.5} & \textbf{69.7 $\pm$ 0.2} & \textbf{75.4 $\pm$ 0.2} & \textbf{49.7 $\pm$ 0.2} & \textbf{46.1 $\pm$ 0.3} \\
\cline{1-8}
\multirow[t]{4}{*}{FA} & Baseline & 49.0 $\pm$ 1.9 & 10.2 $\pm$ 0.4 & 13.2 $\pm$ 1.2 & 1.1 $\pm$ 0.1 & 9.2 $\pm$ 1.5 & 0.5 $\pm$ 0.1 \\
 & LN & 58.1 $\pm$ 1.7 & 38.7 $\pm$ 1.3 & 36.8 $\pm$ 1.2 & 23.7 $\pm$ 0.9 & 29.4 $\pm$ 0.4 & 16.7 $\pm$ 0.2 \\
 & Muon & 70.8 $\pm$ 0.5 & 59.8 $\pm$ 1.1 & \textbf{45.9 $\pm$ 0.5} & 34.7 $\pm$ 0.7 & 31.7 $\pm$ 0.4 & 14.7 $\pm$ 0.4 \\
 & Muon + LN & \textbf{74.3 $\pm$ 0.7} & \textbf{63.8 $\pm$ 0.7} & 45.9 $\pm$ 0.5 & \textbf{43.8 $\pm$ 0.8} & \textbf{33.4 $\pm$ 0.6} & \textbf{21.9 $\pm$ 0.6} \\
\cline{1-8}
\multirow[t]{4}{*}{DFA} & Baseline & 47.3 $\pm$ 1.1 & 11.0 $\pm$ 0.8 & 21.4 $\pm$ 1.5 & 1.2 $\pm$ 0.2 & 12.2 $\pm$ 0.7 & 0.6 $\pm$ 0.1 \\
 & LN & 65.5 $\pm$ 0.3 & 64.5 $\pm$ 0.2 & 44.6 $\pm$ 0.5 & 34.2 $\pm$ 0.3 & 32.1 $\pm$ 0.4 & 24.4 $\pm$ 0.2 \\
 & Muon & 66.6 $\pm$ 1.2 & 60.0 $\pm$ 0.8 & 46.4 $\pm$ 0.3 & 10.3 $\pm$ 0.2 & 27.6 $\pm$ 0.1 & 8.3 $\pm$ 0.6 \\
 & Muon + LN & \textbf{73.3 $\pm$ 0.8} & \textbf{69.7 $\pm$ 0.3} & \textbf{49.1 $\pm$ 0.5} & \textbf{42.3 $\pm$ 0.3} & \textbf{32.5 $\pm$ 0.2} & \textbf{25.1 $\pm$ 0.1} \\
\bottomrule
\end{tabular}
}
\end{table}

\section{Interpolating between SGD and Muon}

\begin{figure}[!htbp]
    \centering
    \subfloat{\includegraphics[width=0.6\linewidth]{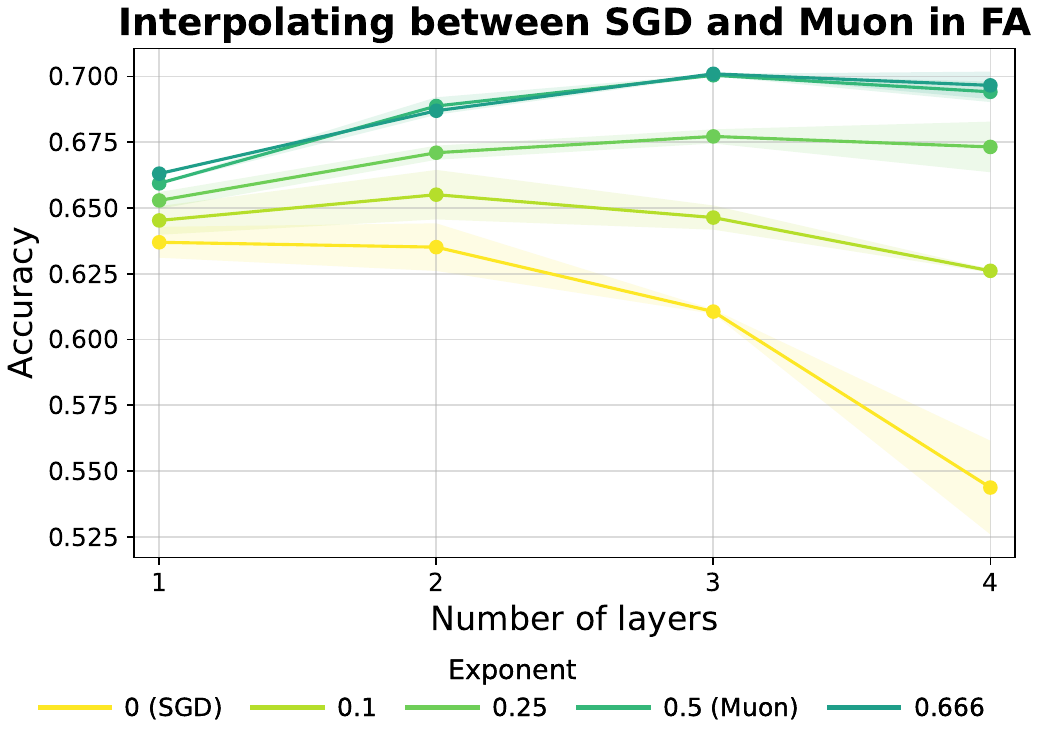}}
    \caption{\textbf{A more balanced singular value spectrum improves performance.} Increasing the exponent in Freon optimiser \citep{shumaylov2026muon}, which, until 0.5, flattens the spectrum of the updates, then makes it skew the other way, increases the performance of the \ac{FA} networks.}
    \label{fig:inter_sgd_muon}
\end{figure}

To better understand how the distribution of the singular values of the updates impacts the performance of the network, we used the Freon optimiser \citep{shumaylov2026muon}, which transforms the momentum matrix $M$ to:  $U \,\operatorname{diag}\!\left(\left(\frac{\sigma}{\|M\|_q}\right)^{q-1}\right) V^\top$, where $U \,\operatorname{diag}\!\left(\sigma\right) V^\top$ is the SVD of $M$. The exponent $c=1-q/2$ controls the transformation of the singular values. With $c=0$, the update is equivalent to SGD with momentum, and directions with large singular values dominate the update. As $c$ increases, the spectrum of the update becomes progressively flatter, giving more weight to directions that would otherwise contribute very little. Using $c=0.5$ corresponds to Muon, where all non-zero singular directions are weighted equally. Using $c>0.5$ scales up the smaller singular values and scales down the bigger ones. This interpolation is relevant for our work since $c$ controls the dimensionality of the updates, complementing the norm-matched rank control in \figref{fig:fig_muon} E. When applied to \ac{FA} models, performance improves as the Freon exponent is increased from the SGD regime toward the Muon regime (\figref{fig:inter_sgd_muon}), supporting the claim that making the update higher-dimensional improves \ac{FA}.

\section{Theoretical analysis of Feedback Alignment}    
\label{sec:fa_theory}
In this section, we first explain when imperfect alignment may slow down optimisation, then we provide intuition for why the singular values can become more concentrated across layers when training with \ac{FA}. By doing so, we bring new perspectives to past theoretical analysis of \ac{FA} \citep{lillicrap2016random,girotti2021convergence,robertson2023implicit}. 
Much of the theory on neural network learning dynamics is done in the analytically tractable setting of linear networks \citep{fukumizu1998effect,saxe2013exact,arora2018optimization,arora2019implicit}. 
We follow this setup, and consider a linear network without biases, with weights $W_\ell\in\mathbb{R}^{n_\ell\times n_{\ell-1}}$ and end-to-end weights $W_e=W_L\cdots W_1$. Throughout this section, we use gradient flow with a learning rate $\eta>0$, and no momentum or weight decay. 
We write $G=\nabla_{W_e}E(W_e)$ for the gradient of the loss with respect to the end-to-end weights.

First, for \ac{BP}, \citet{arora2018optimization} showed that depth changes the optimisation dynamics of linear networks by preconditioning the end-to-end gradient:
\begin{equation}
\dot{W}_e^{BP}
=-\eta\sum_{j=1}^{L}
\left[W_e^{BP}(W_e^{BP})^\top\right]^{\frac{j-1}{L}}
G
\left[(W_e^{BP})^\top W_e^{BP}\right]^{\frac{L-j}{L}}.
\label{eq:bp_impl}
\end{equation}
This results from the layers of the network being balanced, which means that if the weights are initialised such that $W_jW_j^\top=W_{j+1}^\top W_{j+1}$  then it holds throughout training.  The preconditioner depends on the singular values and singular vectors of the end-to-end weights. This relation between the geometry of the weights and the optimisation dynamics is central to studies of implicit regularisation in deep linear networks \citep{arora2019implicit,ji2018gradient}.

To obtain the \ac{FA} end-to-end dynamics, we define the forward prefix $A_\ell=W_{\ell-1}\cdots W_1$, the forward suffix $R_\ell=W_L\cdots W_{\ell+1}$, and the cumulative feedback map $F_\ell=B_{\ell+1}\cdots B_L$, with $A_1=I_{n_0}$ and $R_L=F_L=I_{n_L}$. For linear activations, Eq.~\ref{eq:fa_rec} gives $\dot W_\ell=-\eta F_\ell G A_\ell^\top$. Applying the product rule, $\dot W_e^{FA}=\sum_{\ell=1}^L R_\ell\dot W_\ell A_\ell$, therefore gives:
\begin{equation}
\dot{W}_e^{FA}
=-\eta\sum_{\ell=1}^{L}(R_\ell F_\ell)G(A_\ell^\top A_\ell).
\label{eq:fa_impl}
\end{equation}
Writing Eq.~\ref{eq:fa_impl} in vectorised form, with $g=\operatorname{vec}(G)$ and columns stacked in $\operatorname{vec}$, gives:
\begin{equation}
\operatorname{vec}(\dot W_e^{FA})=-\eta\mathcal{P}g,
\qquad
\mathcal{P}=\sum_{\ell=1}^{L}(A_\ell^\top A_\ell)\otimes(R_\ell F_\ell).
\label{eq:fa_impl_vec}
\end{equation}
Unlike the preconditioning induced by \ac{BP}, $\mathcal{P}$ is not necessarily positive semidefinite, therefore the preconditioning may not keep the loss decreasing. To see that, we can compute the change in loss along this flow, by applying the chain rule to $E(W_e(t))$:
\[
\dot E
=\operatorname{tr}(G^\top\dot W_e^{FA})
=g^\top\operatorname{vec}(\dot W_e^{FA})
=-\eta g^\top\mathcal{P}g,
\]
where the last equality uses Eq.~\ref{eq:fa_impl_vec}. Since $g^\top\mathcal{P}g$ is a scalar, it equals its transpose $g^\top\mathcal{P}^\top g$. Averaging the two expressions gives
\[
g^\top\mathcal{P}g
=\frac{1}{2}\left(g^\top\mathcal{P}g+g^\top\mathcal{P}^\top g\right)
=g^\top\operatorname{sym}(\mathcal{P})g,
\]
with $\operatorname{sym}(\mathcal{P})=(\mathcal{P}+\mathcal{P}^\top)/2$. Thus, only the symmetric part of $\mathcal{P}$ contributes to the instantaneous change in loss. The loss therefore decreases whenever $g^\top\operatorname{sym}(\mathcal{P})g>0$. Positive semidefiniteness of $\operatorname{sym}(\mathcal{P})$ guarantees that the loss does not increase for any gradient, but is not necessary for a decrease along the current gradient. Still, poor alignment can make the quadratic form small or negative, resulting in slow progress or an increase in loss. 

To understand how the weight spectra can change across layers in \ac{FA}, we first explain that singular values are shared across layers in \ac{BP}. We then derive an identity related to the balanced-layer invariant in \ac{BP} \citep{arora2018optimization,du2018algorithmic}. This identity was used by \citet{lillicrap2016random} for two-layer linear networks. A related neuronwise conservation law for nonlinear networks was studied by \citet{robertson2023implicit}.

For balanced linear \ac{BP}, the relation $W_\ell W_\ell^\top=W_{\ell+1}^\top W_{\ell+1}$ is preserved by gradient flow when it holds at initialisation. The nonzero eigenvalues of these Gram matrices are the squared nonzero singular values of the corresponding weights, so adjacent layers have identical nonzero singular values: $\sigma_i(W_\ell)=\sigma_i(W_{\ell+1})$ for $i=1,\ldots,r$, where $r$ is their common rank. Thus, whenever $r>0$, the ratios $\sigma_i(W_\ell)/\sigma_1(W_\ell)$ are the same across layers, although the shared spectrum can evolve during training.

For a linear network trained by \ac{FA} gradient flow with fixed feedback weights, the quantity
\begin{equation}
C_\ell=W_\ell W_\ell^\top-B_{\ell+1}W_{\ell+1}-W_{\ell+1}^\top B_{\ell+1}^\top
\label{eq:fa_balanced}
\end{equation}
is conserved for every adjacent pair of layers, $\ell=1,\ldots,L-1$. In particular, $C_\ell(t)=C_\ell(0)$ throughout training.

\paragraph{Proof.}
The layer dynamics are $\dot W_\ell=-\eta F_\ell G A_\ell^\top$. Using $F_\ell=B_{\ell+1}F_{\ell+1}$ and $A_{\ell+1}=W_\ell A_\ell$, we obtain:
\begin{equation}
\begin{aligned}
\dot W_\ell W_\ell^\top
&=-\eta B_{\ell+1}F_{\ell+1}GA_\ell^\top W_\ell^\top \\
&=-\eta B_{\ell+1}F_{\ell+1}GA_{\ell+1}^\top
=B_{\ell+1}\dot W_{\ell+1}.
\end{aligned}
\label{eq:fa_invariant_derivative}
\end{equation}
Since $B_{\ell+1}$ is fixed, differentiating Eq.~\ref{eq:fa_balanced} and applying this relation and its transpose gives:
\[
\begin{aligned}
\dot C_\ell
&=\dot W_\ell W_\ell^\top+W_\ell\dot W_\ell^\top
-B_{\ell+1}\dot W_{\ell+1}-\dot W_{\ell+1}^\top B_{\ell+1}^\top  =0.
\end{aligned}
\]
Hence $C_\ell$ remains equal to its initial value.\hfill$\square$

\paragraph{General relation between layer spectra.}
Let $W_{\ell+1}v_i=\tau_i u_i$ be a singular pair of the upper layer, with $\tau_i>0$ and unit vectors $u_i,v_i$. Write $s_j=\sigma_j(W_\ell)$ and let $p_j$ be an orthonormal eigenbasis of $W_\ell W_\ell^\top$. Define the directional feedback gain $\alpha_i=v_i^\top B_{\ell+1}u_i$ and the initialisation contribution $c_i=v_i^\top C_\ell v_i$. The invariant gives the exact relation:
\begin{equation}
2\alpha_i\tau_i
=\|W_\ell^\top v_i\|_2^2-c_i
=\sum_{j=1}^{n_\ell}s_j^2|\langle p_j,v_i\rangle|^2-c_i.
\label{eq:fa_spectral_general}
\end{equation}
Thus, the upper-layer singular values depend on the squared lower-layer spectrum, the overlap between neighbouring singular directions, and the feedback gain along each direction.

\paragraph{Proof.}
Multiplying Eq.~\ref{eq:fa_balanced} on the left by $v_i^\top$ and on the right by $v_i$ gives:
\[
\begin{aligned}
\|W_\ell^\top v_i\|_2^2-c_i
&=v_i^\top B_{\ell+1}W_{\ell+1}v_i
+v_i^\top W_{\ell+1}^\top B_{\ell+1}^\top v_i \\
&=2\tau_i v_i^\top B_{\ell+1}u_i
=2\alpha_i\tau_i.
\end{aligned}
\]
Expanding $v_i$ in the basis $(p_j)$ gives the sum in Eq.~\ref{eq:fa_spectral_general}.\hfill$\square$

An identity on relative singular values follows by controlling the directional quantities above. Order the singular values of both layers from largest to smallest, assume $s_1>0$, and define
\begin{equation}
e_i=\|W_\ell^\top v_i\|_2^2-s_i^2-c_i.
\label{eq:fa_spectral_residual}
\end{equation}
This residual includes both the mixing of lower-layer modes and the contribution of the conserved initialisation term. Eq.~\ref{eq:fa_spectral_general} gives:
\begin{equation}
\tau_i=\frac{s_i^2+e_i}{2\alpha_i}
\label{eq:fa_spectral_ratio_bound}
\end{equation}

If the residuals vanish and the gains are equal, the ratios are squared exactly, as illustrated by the following special case.

\paragraph{Spectral squaring across layers.}
As an example of Eq.~\ref{eq:fa_spectral_general}, consider square layers of equal width and scaled-orthogonal feedback, $B_{\ell+1}=b_{\ell+1}Q_{\ell+1}$, where $b_{\ell+1}>0$ and $Q_{\ell+1}^\top Q_{\ell+1}=I$. Keeping the notation $s_i=\sigma_i(W_\ell)$ and $\tau_i=\sigma_i(W_{\ell+1})$, ordered from largest to smallest. Assuming that $C_\ell=0$ and $K_\ell=\operatorname{skew}(Q_{\ell+1}W_{\ell+1})=0$, where $\operatorname{skew}(X)=(X-X^\top)/2$. Then:
\begin{equation}
\tau_i=\frac{s_i^2}{2b_{\ell+1}},
\qquad
\frac{\tau_i}{\tau_1}
=\left(\frac{s_i}{s_1}\right)^2.
\label{eq:fa_spectral_squaring}
\end{equation}
where the ratio assumes $s_1>0$. Thus, a mode with $0<s_i/s_1<1$ becomes relatively smaller in the next layer. Using recursive application across layers, write $s_i^{(k)}=\sigma_i(W_k)$, so that $s_i=s_i^{(\ell)}$ and $\tau_i=s_i^{(\ell+1)}$. If these conditions hold at every layer, then
\begin{equation}
\frac{s_i^{(k)}}{s_1^{(k)}}
=\left(\frac{s_i^{(1)}}{s_1^{(1)}}\right)^{2^{k-1}}.
\label{eq:fa_spectral_depth}
\end{equation}

\paragraph{Proof.}
Eq.~\ref{eq:fa_balanced} can be rearranged as:
\begin{equation}
Q_{\ell+1}W_{\ell+1}
=\frac{W_\ell W_\ell^\top-C_\ell}{2b_{\ell+1}}+K_\ell,
\qquad
K_\ell=\operatorname{skew}(Q_{\ell+1}W_{\ell+1}).
\label{eq:fa_spectral_decomposition}
\end{equation}
The invariant therefore determines the symmetric part of $Q_{\ell+1}W_{\ell+1}$; its skew-symmetric part $K_\ell$ can still change during training. When $C_\ell=0$ and $K_\ell=0$, this reduces to $Q_{\ell+1}W_{\ell+1}=W_\ell W_\ell^\top/(2b_{\ell+1})$. Using the eigenbasis $(p_i)$ from the general relation, we can therefore choose $v_i=p_i$ and $u_i=Q_{\ell+1}^\top p_i$, which gives
\[
W_{\ell+1}v_i=\frac{s_i^2}{2b_{\ell+1}}u_i,
\qquad
\alpha_i=v_i^\top B_{\ell+1}u_i=b_{\ell+1},
\qquad
c_i=e_i=0.
\]
Thus, for each positive mode, Eq.~\ref{eq:fa_spectral_general} reduces to $2b_{\ell+1}\tau_i=s_i^2$. Dividing by the leading singular value gives the squared-ratio relation in Eq.~\ref{eq:fa_spectral_squaring}. Application across layers gives Eq.~\ref{eq:fa_spectral_depth}.\hfill$\square$

A similar relation was found in \citep{girotti2021convergence} for diagonal linear \ac{FA} dynamics, where the feedback is fixed and the weights are initialised to be diagonal. In that setting, the singular values of each layer are simply the absolute values of the diagonal entries, and the squared-ratio relation holds exactly. The more general relation in Eq.~\ref{eq:fa_spectral_general} shows that this behaviour can also emerge in non-diagonal networks, provided that the directional feedback gains are similar across modes and that the residuals are small.

These results identify a mechanism by which FA can make optimisation harder than BP. 
In balanced linear BP, adjacent layers have identical nonzero singular values, preserving their relative spectra across depth. 
In FA, the invariant instead couples upper-layer singular values to mixtures of squared lower-layer singular values, modulated by directional feedback gains and the conserved initialisation term (Eq.~\ref{eq:fa_spectral_general}). 
Under the spectral-squaring conditions, successive layers increasingly separate weak modes from dominant ones (Eq.~\ref{eq:fa_spectral_depth}), consistent with related power relations in diagonal linear FA dynamics \citep{girotti2021convergence}. 
 Task-relevant errors in these directions can consequently receive much smaller corrective updates than errors in dominant directions, even when the overall update decreases the loss. BP can also exhibit unequal learning rates across directions, but its end-to-end preconditioner is positive semidefinite. FA lacks this guarantee, allowing poor alignment to further reduce progress or even increase the loss. 
 Spectral concentration can therefore compound imperfect feedback alignment, slowing the optimisation of task-relevant errors outside the dominant learning directions.

\section{Additional metrics}

For each panel, every seed is summarised by the mean of the metric over the last 20\% of training. 
Each group is then compared with its reference at the same depth using a two-sided Welch's $t$-test (5 seeds per group). 
The references are \ac{BP} for the feedback methods, AdamW for Muon, and no normalisation for \ac{BN}. Muon and \ac{BN} are compared with their reference separately for each feedback method. 
Every group uses the hyperparameter configuration with the highest mean test accuracy for that depth. $p$-values are Holm-Bonferroni corrected across all comparisons in a figure. Symbols to the right of each panel give the corrected significance (*~$p<0.05$, **~$p<0.01$, ***~$p<0.001$, n.s.~not significant), and arrows show whether the group lies above ($\uparrow$) or below ($\downarrow$) its reference.

\subsection{Comparing feedback methods without Muon or Batch Norm}

\begin{figure}[!htbp]
    \centering
    \subfloat{\includegraphics[width=0.76\linewidth]{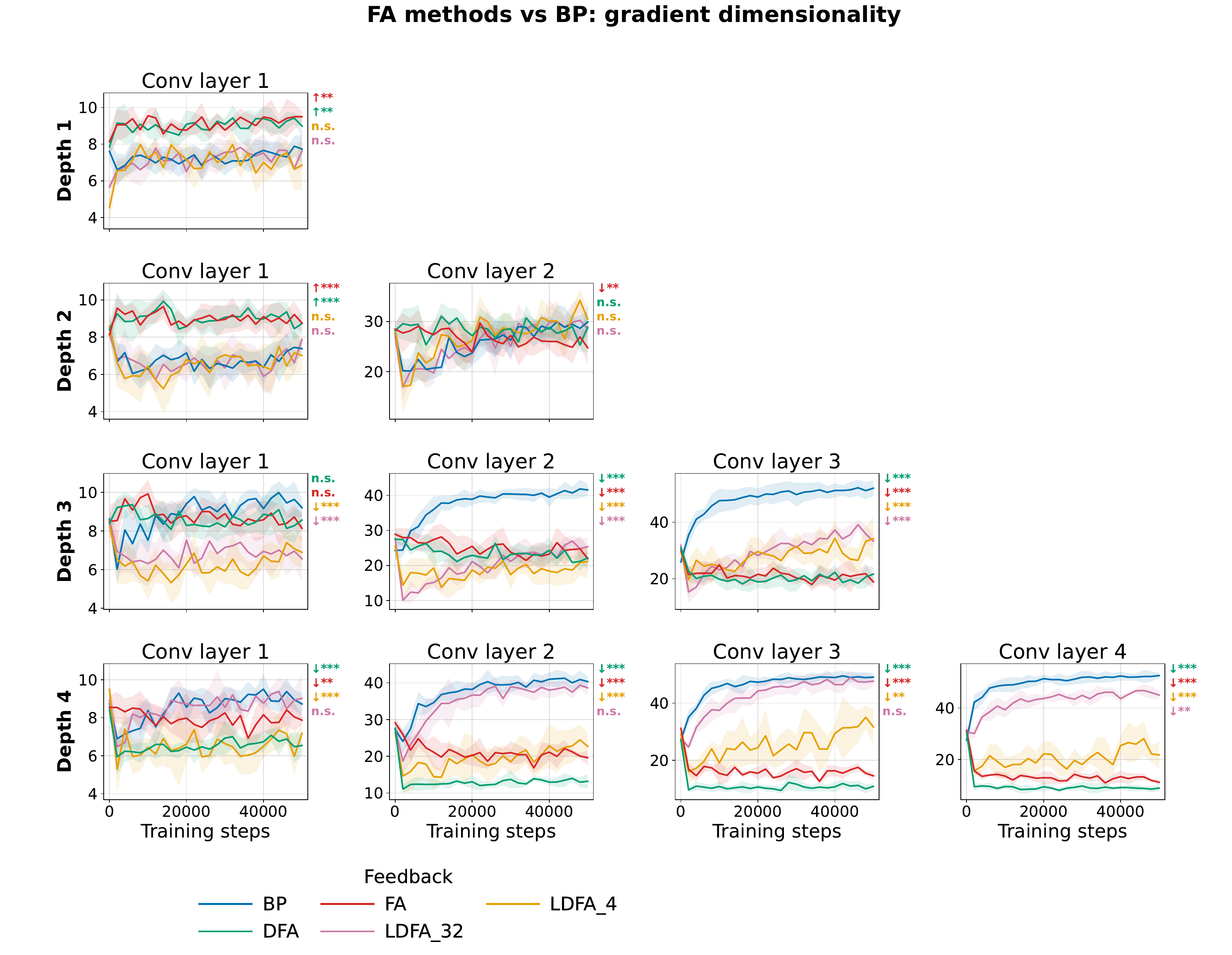}}
    \caption{\textbf{\ac{FA} and \ac{DFA} have lower gradient dimensionality than \ac{BP} in deep networks.} Effective rank of the gradients over training for each layer (columns) and network depth (rows), all trained with AdamW and without normalisation. Lines show the mean over 5 seeds and shading $\pm1$ standard deviation.}
    \label{fig:app_bp_fa_grad_rank}
\end{figure}

\begin{figure}[!htbp]
    \centering
    \subfloat{\includegraphics[width=0.76\linewidth]{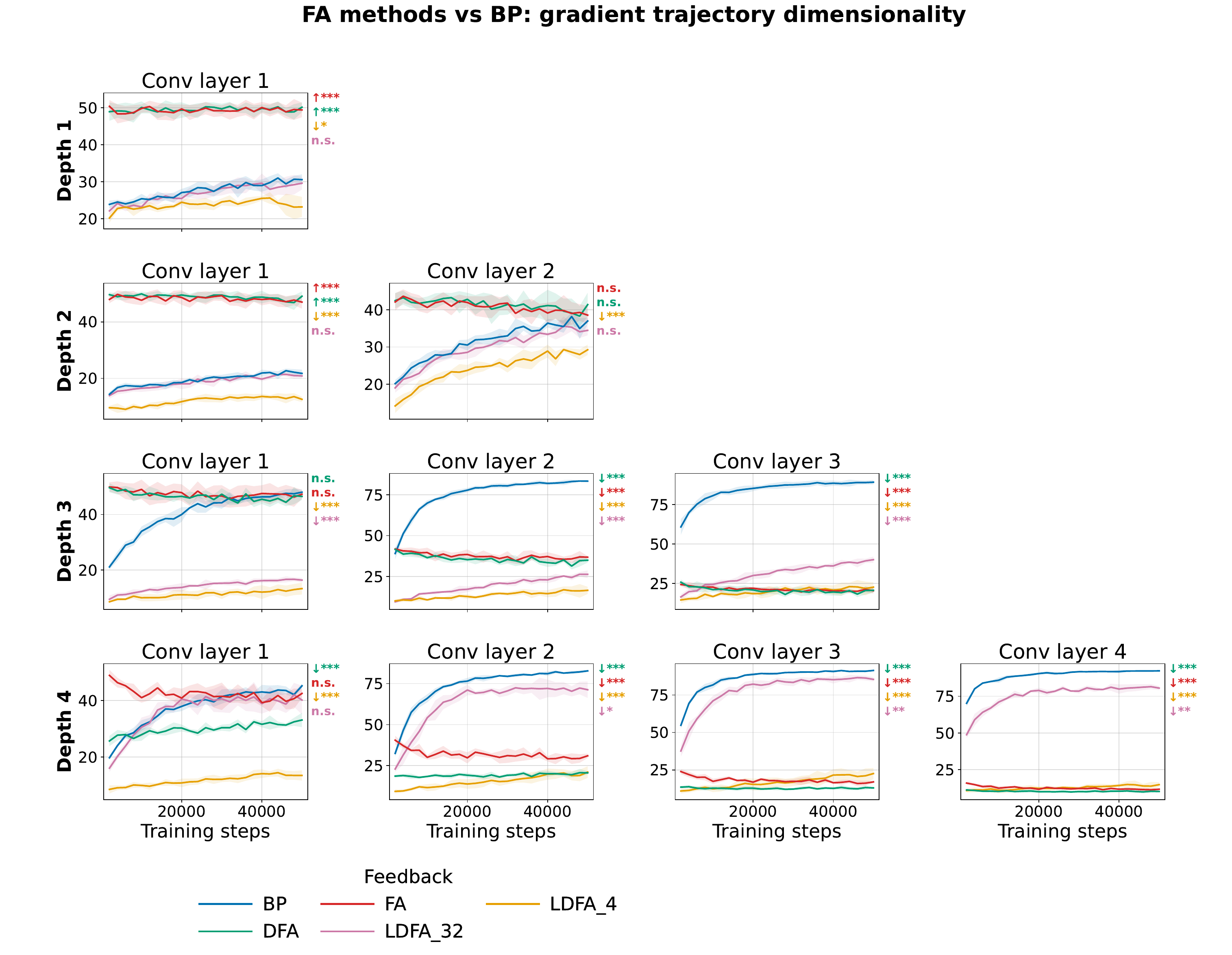}}
    \caption{\textbf{\ac{FA} and \ac{DFA} have lower-dimensional gradient trajectories than \ac{BP} in deep networks.} Effective rank of the Gram matrix of the gradient trajectory for each layer (columns) and network depth (rows), all trained with AdamW and without normalisation. Lines show the mean over 5 seeds and shading $\pm1$ standard deviation.}
    \label{fig:app_bp_fa_grad_traj_rank}
\end{figure}

In deep networks, \ac{FA} and \ac{DFA} produce gradients with significantly lower effective rank than \ac{BP} in almost every layer (\figref{fig:app_bp_fa_grad_rank}). 
The gap widens with layer index: in four layer networks, the reduction grows from 12\% in the first layer to 76\% in the last layer for \ac{FA}, and from 25\% to 83\% for \ac{DFA}. \ac{LDFA} with rank 4 shows a weaker but still significant reduction.
 \ac{LDFA} with rank 32 stays close to \ac{BP}, differing significantly only in the last layer. The deficit appears with depth: in one- and two-layer networks, \ac{FA} and \ac{DFA} actually have a significantly higher first-layer gradient rank than \ac{BP}. 
 Gradient trajectories show the same pattern (\figref{fig:app_bp_fa_grad_traj_rank}). In four layer networks, \ac{FA} and \ac{DFA} trajectories are 64-89\% lower-dimensional than those of \ac{BP} beyond the first layer. \ac{LDFA} with rank 4 is significantly lower at every depth.

\subsection{Comparing Muon with AdamW}
\begin{figure}[!htbp]
    \centering
    \subfloat{\includegraphics[width=0.76\linewidth]{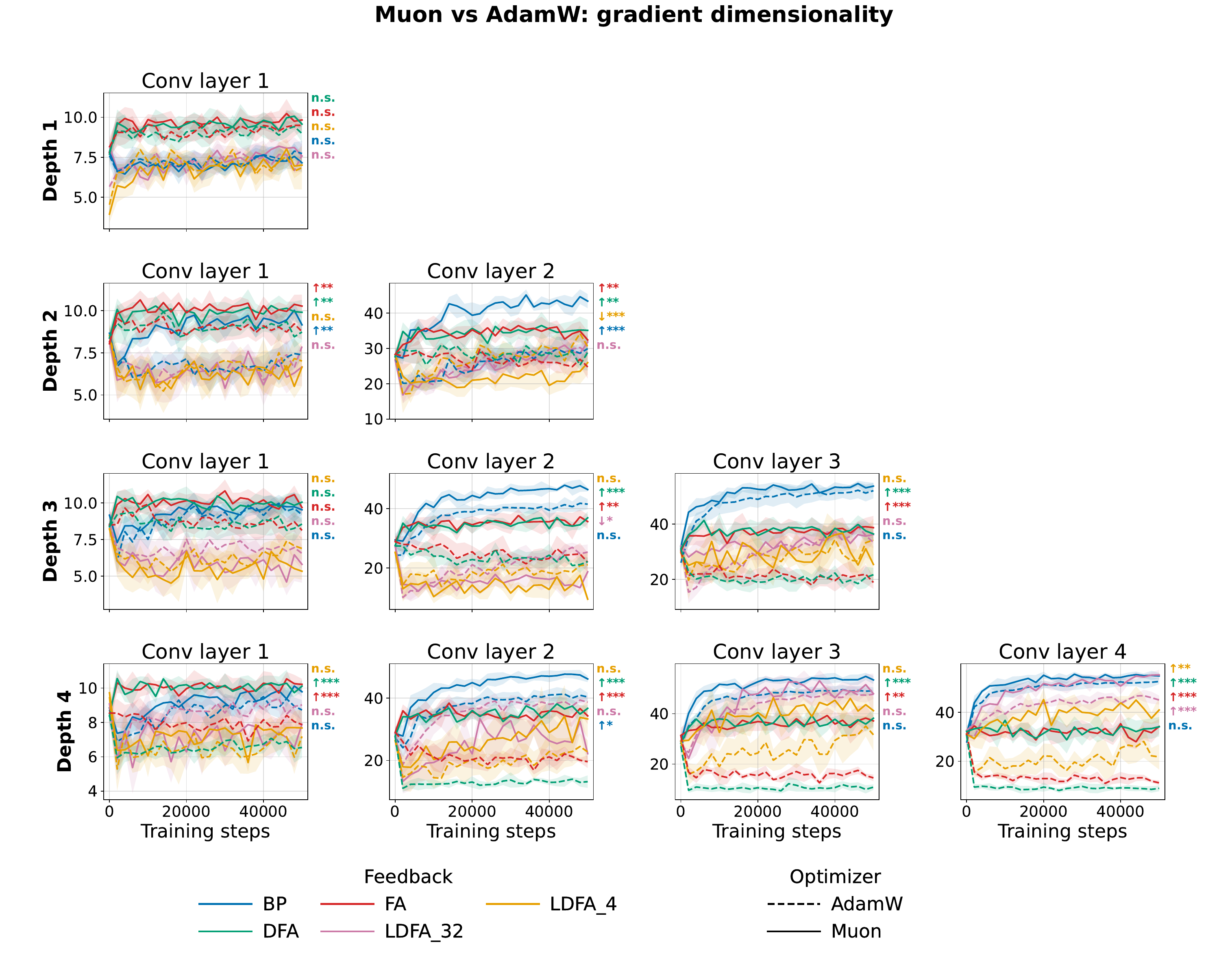}}
    \caption{\textbf{Muon increases the gradient dimensionality of \ac{FA} and \ac{DFA} across layers.} Effective rank of the gradients for each layer (columns) and network depth (rows), for every feedback method trained with Muon (solid) or AdamW (dashed). Lines show the mean over 5 seeds and shading $\pm1$ standard deviation.}
    \label{fig:app_muon_grad_rank}
\end{figure}

\begin{figure}[!htbp]
    \centering
    \subfloat{\includegraphics[width=0.76\linewidth]{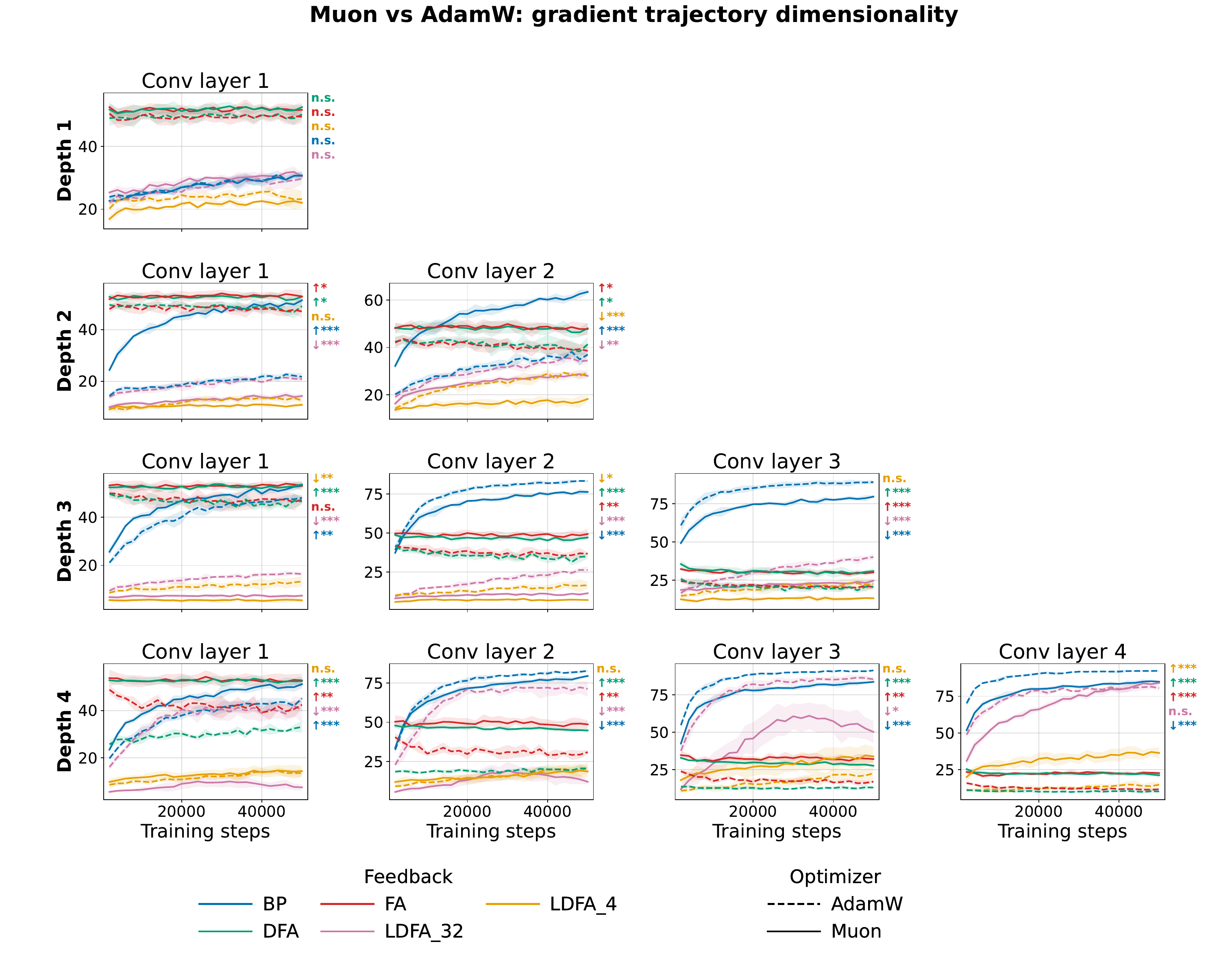}}
    \caption{\textbf{Muon increases the gradient trajectory dimensionality of \ac{FA} and \ac{DFA}} Effective rank of the Gram matrix of the gradient trajectory for each layer (columns) and network depth (rows), for every feedback method trained with Muon (solid) or AdamW (dashed). Lines show the mean over 5 seeds and shading $\pm1$ standard deviation.}
    \label{fig:app_muon_grad_traj_rank}
\end{figure}

Muon significantly increases the gradient effective rank of \ac{FA} and \ac{DFA} networks in every layer of four layer networks (\figref{fig:app_muon_grad_rank}), by 28-154\% for \ac{FA} and 50-277\% for \ac{DFA}. 
The effect is largest in the deepest layers, where AdamW-trained networks are most collapsed. In contrast, Muon barely changes the gradient rank of \ac{BP} (5-16\%, mostly not significant).
 For \ac{LDFA}, it only raises the rank significantly in the last layer. The trajectory dimensionality of \ac{FA} and \ac{DFA} also increases significantly (\figref{fig:app_muon_grad_traj_rank}; 29-94\% and 63-124\%). 
 For \ac{BP}, Muon slightly but significantly lowers trajectory dimensionality in the deeper layers. For \ac{LDFA} with rank 32, it lowers it in most layers. The benefit of Muon is therefore specific to methods whose updates are low-dimensional under AdamW.

\subsection{Comparing Batch Norm with no normalisation}

\begin{figure}[!htbp]
    \centering
    \subfloat{\includegraphics[width=0.76\linewidth]{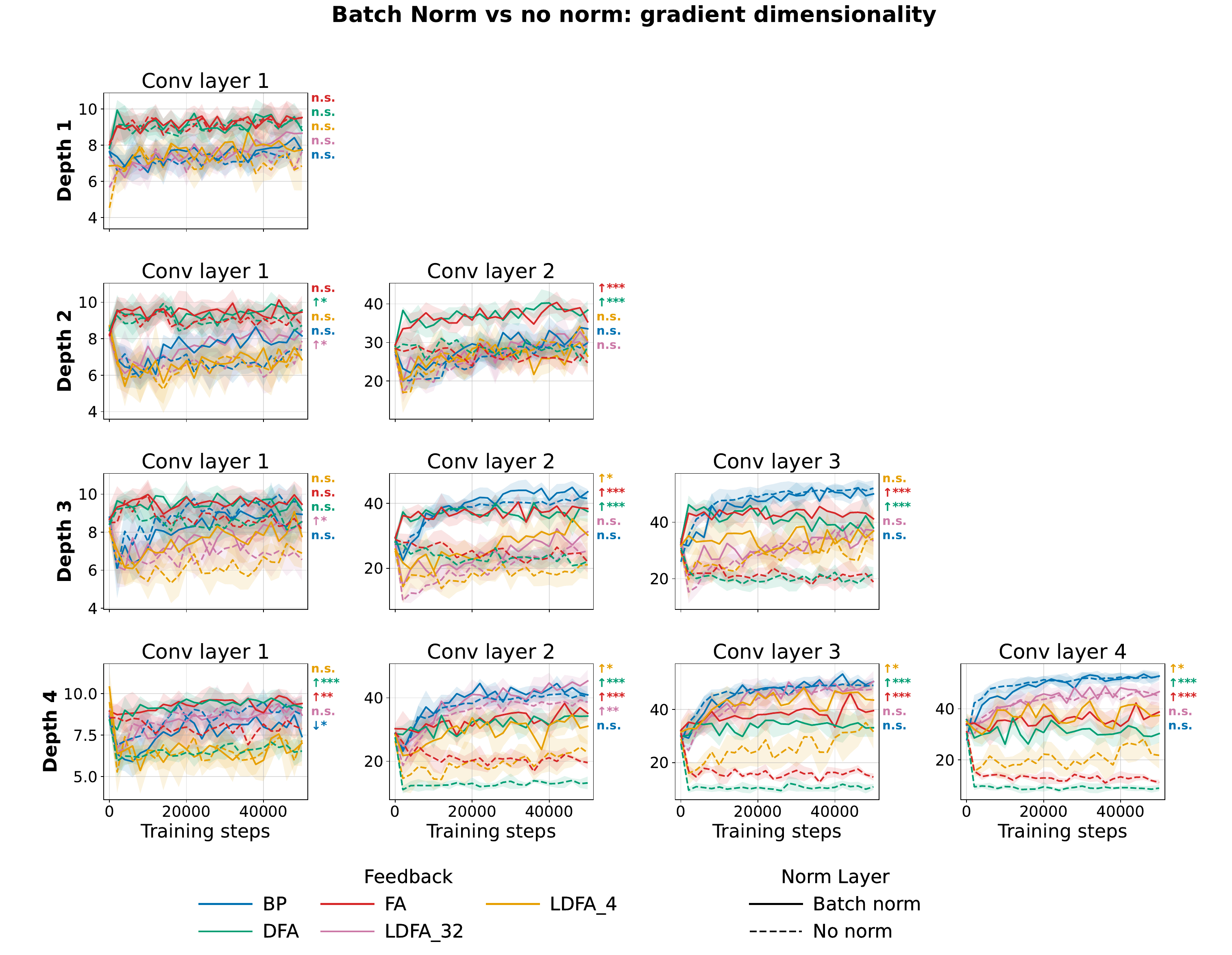}}
    \caption{\textbf{\ac{BN} increases the gradient dimensionality of \ac{FA} and \ac{DFA} across layers.} Effective rank of the gradients for each layer (columns) and network depth (rows), for every feedback method trained with AdamW, with \ac{BN} (solid) or without normalisation (dashed). Lines show the mean over 5 seeds and shading $\pm1$ standard deviation. }
    \label{fig:app_bnorm_grad_rank}
\end{figure}

\begin{figure}[!htbp]
    \centering
    \subfloat{\includegraphics[width=0.76\linewidth]{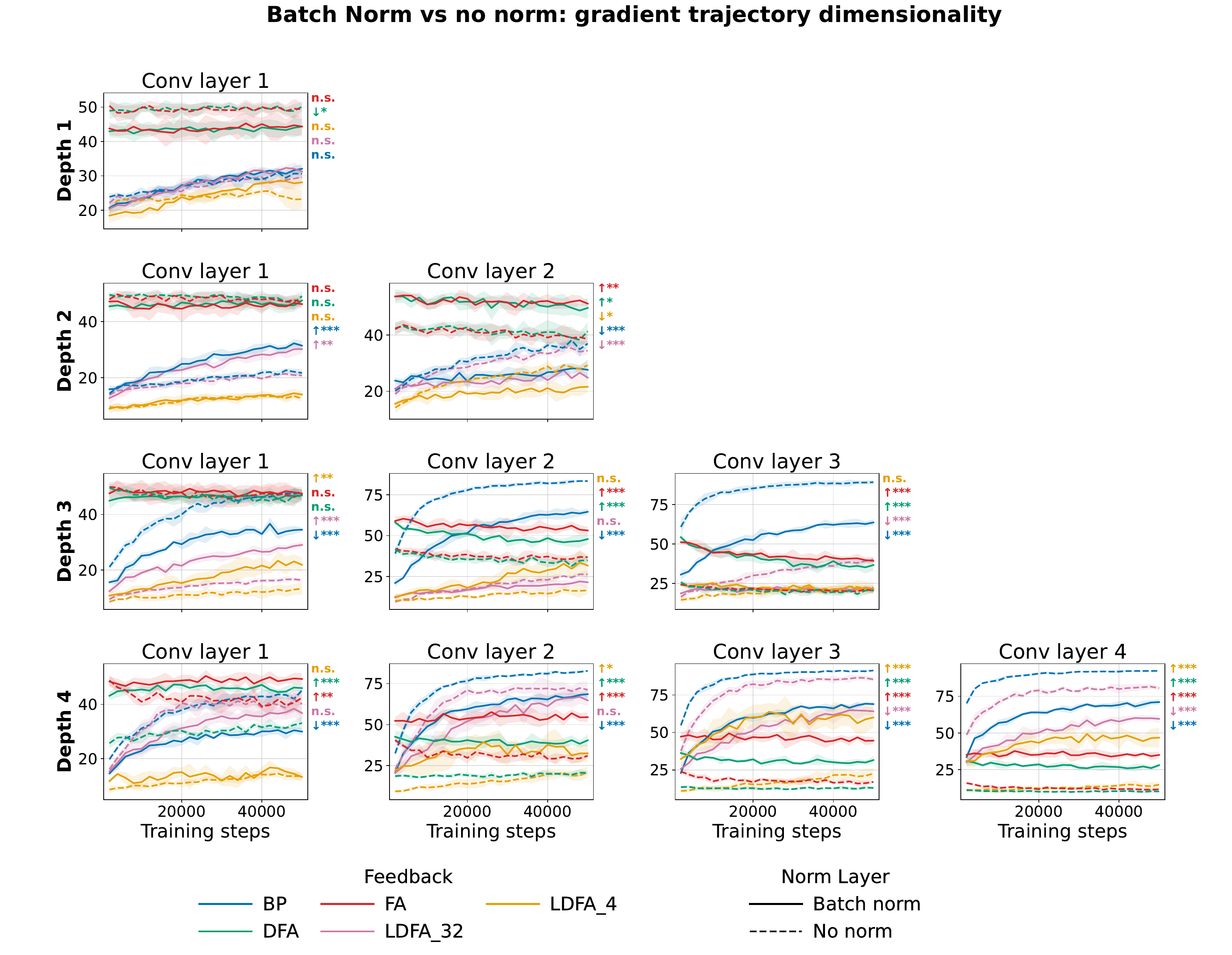}}
    \caption{\textbf{\ac{BN} increases the gradient trajectory dimensionality of \ac{FA} and \ac{DFA}.} Effective rank of the Gram matrix of the gradient trajectory for each layer (columns) and network depth (rows), for every feedback method trained with AdamW, with \ac{BN} (solid) or without normalisation (dashed). Lines show the mean over 5 seeds and shading $\pm1$ standard deviation.}
    \label{fig:app_bnorm_grad_traj_rank}
\end{figure}

\ac{BN} has an effect similar to Muon. It significantly increases the gradient effective rank of \ac{FA} and \ac{DFA} networks at depths three and four (\figref{fig:app_bnorm_grad_rank}). 
In four layer networks, the increase grows from 18\% to 196\% across layers for \ac{FA} and from 39\% to 249\% for \ac{DFA}. 
\ac{BP} is essentially unaffected. \ac{LDFA} with rank 4 shows a smaller significant increase (41-59\%), and rank 32 is mostly unchanged. 
\ac{BN} also significantly increases the trajectory dimensionality of \ac{FA} and \ac{DFA} (\figref{fig:app_bnorm_grad_traj_rank}), by up to 201\% and 160\%. For \ac{BP}, it reduces trajectory dimensionality by 18-31\%. 
Like Muon, \ac{BN} therefore mainly helps the methods whose gradients collapse without it.

\section{Reproducibility}
For sections ~\ref{sec:patho} and ~\ref{sec:explo}, we performed hyperparameter searches with 2 seeds,  then selected the best validation hyperparameters and ran the simulations for 5 seeds. We used the cross-entropy loss, mini-batch size 64 for 50,000 steps, the configured optimiser, while periodically evaluating train/test loss, and gradient-trajectory metrics at fixed intervals with early stopping for convergence or failed runs. 

For unbiased results, for every feedback method, we optimised each models' hyperparameter separately using a grid search over network depths (1,2,3,4), normalisation in (None, Batch Normalisation), Muon with learning rates $({10^{-1},10^{-2},10^{-3},10^{-4},10^{-5}})$ and momentum $({0,0.9,0.95})$, and weight decay $({0,10^{-3}})$, and AdamW with learning rates $({10^{-2},10^{-3},10^{-4},10^{-5}})$ and weight decay $({0,10^{-3}})$.

For section~\ref{sec:scaling}, we used the epoch-based training pipeline for vision models on STL-10, CIFAR-100, and ImageNet. STL-10 and CIFAR-100 models were trained for 90 epochs with a batch size of 128; hyperparameters were searched with one seed and the selected configurations were evaluated with five seeds. ImageNet models were trained for 50 epochs with a batch size of 256 and 3 seeds. We used automatic mixed precision with float16 and a cosine-annealing learning-rate schedule.

We selected hyperparameters separately for each dataset, architecture, feedback method (\ac{BP}, \ac{FA}, or \ac{DFA}), and normalisation setting. ImageNet used configuration-specific learning-rate searches, rather than the grid used for the smaller-scale experiments. ImageNet training used random resized crops to $224\times224$, random horizontal flips, and RandAugment; evaluation used resizing to 256 followed by a $224\times224$ centre crop. We used the standard training and test splits for STL-10 and CIFAR-100, and the training and validation splits for ImageNet. Table~\ref{tab:scale} reports the mean and standard deviation across available seeds of each run's best top-1 evaluation accuracy.

Experiments were run in PyTorch \citep{paszke2019pytorch} on a workstation equipped with two NVIDIA RTX PRO 4000 Blackwell GPUs, with Hydra \citep{Yadan2019Hydra} used to manage model, optimiser, data, and training hyperparameters across grid-search runs. The resulting accuracy and loss curves were visualised using Matplotlib \citep{Hunter:2007}. One full run on the CIFAR10 dataset with 50,000 steps took approximately 10 minutes. One full run on STL-10 took approximately 10 minutes.

\clearpage

\section*{NeurIPS Paper Checklist}

\begin{enumerate}

\item {\bf Claims}
    \item[] Question: Do the main claims made in the abstract and introduction accurately reflect the paper's contributions and scope?
    \item[] Answer: \answerYes{}{} 
    \item[] Justification: In the abstract and introduction, we summarized our work and explained its contributions, and did not claim anything further.

\item {\bf Limitations}
    \item[] Question: Does the paper discuss the limitations of the work performed by the authors?
    \item[] Answer: \answerYes{} 
    \item[] Justification: We discussed the limitations in the discussion section.

\item {\bf Theory assumptions and proofs}
    \item[] Question: For each theoretical result, does the paper provide the full set of assumptions and a complete (and correct) proof?
    \item[] Answer: \answerYes{} 
    \item[] Justification: We provide the full set of assumptions and proofs in the appendix.

    \item {\bf Experimental result reproducibility}
    \item[] Question: Does the paper fully disclose all the information needed to reproduce the main experimental results of the paper to the extent that it affects the main claims and/or conclusions of the paper (regardless of whether the code and data are provided or not)?
    \item[] Answer: \answerYes{} 
    \item[] Justification: We put the information in the appendix

\item {\bf Open access to data and code}
    \item[] Question: Does the paper provide open access to the data and code, with sufficient instructions to faithfully reproduce the main experimental results, as described in supplemental material?
    \item[] Answer: \answerYes{} 
    \item[] Justification: We provided anonymized code in the supplemental material.

\item {\bf Experimental setting/details}
    \item[] Question: Does the paper specify all the training and test details (e.g., data splits, hyperparameters, how they were chosen, type of optimizer) necessary to understand the results?
    \item[] Answer: \answerYes{} 
    \item[] Justification: Yes, we specified all of these details in the appendix.

\item {\bf Experiment statistical significance}
    \item[] Question: Does the paper report error bars suitably and correctly defined or other appropriate information about the statistical significance of the experiments?
    \item[] Answer: \answerYes{} 
    \item[] Justification: We used 5 random seeds for each result, except for ImageNet where we used 3 seeds.

\item {\bf Experiments compute resources}
    \item[] Question: For each experiment, does the paper provide sufficient information on the computer resources (type of compute workers, memory, time of execution) needed to reproduce the experiments?
    \item[] Answer: \answerYes{} 
    \item[] Justification: We provided this information in the appendix.
    
\item {\bf Code of ethics}
    \item[] Question: Does the research conducted in the paper conform, in every respect, with the NeurIPS Code of Ethics \url{https://neurips.cc/public/EthicsGuidelines}?
    \item[] Answer: \answerYes{} 
    \item[] Justification: We followed strictly the code of ethics.

\item {\bf Broader impacts}
    \item[] Question: Does the paper discuss both potential positive societal impacts and negative societal impacts of the work performed?
    \item[] Answer: \answerNA{} 
    \item[] Justification: We do not expect any societal impact from this work. It is mainly about understanding alternatives to backpropagation, which is useful for neuroscience, but will not currently advance machine learning capabilities.
    
\item {\bf Safeguards}
    \item[] Question: Does the paper describe safeguards that have been put in place for responsible release of data or models that have a high risk for misuse (e.g., pre-trained language models, image generators, or scraped datasets)?
    \item[] Answer: \answerNA{} 
    \item[] Justification: Our work does not pose any risk of misuse.

\item {\bf Licenses for existing assets}
    \item[] Question: Are the creators or original owners of assets (e.g., code, data, models), used in the paper, properly credited and are the license and terms of use explicitly mentioned and properly respected?
    \item[] Answer: \answerYes{} 
    \item[] Justification: We cited appropriately previous work, and code we used.  

\item {\bf New assets}
    \item[] Question: Are new assets introduced in the paper well documented and is the documentation provided alongside the assets?
    \item[] Answer: \answerNA{} 
    \item[] Justification: We did not release new assets.

\item {\bf Crowdsourcing and research with human subjects}
    \item[] Question: For crowdsourcing experiments and research with human subjects, does the paper include the full text of instructions given to participants and screenshots, if applicable, as well as details about compensation (if any)? 
    \item[] Answer: \answerNA{} 
    \item[] Justification: Our work did not involve crowdsourcing experiments.

\item {\bf Institutional review board (IRB) approvals or equivalent for research with human subjects}
    \item[] Question: Does the paper describe potential risks incurred by study participants, whether such risks were disclosed to the subjects, and whether Institutional Review Board (IRB) approvals (or an equivalent approval/review based on the requirements of your country or institution) were obtained?
    \item[] Answer: \answerNA{} 
    \item[] Justification:  Our work did not involve crowdsourcing experiments nor research with human subjects.

\item {\bf Declaration of LLM usage}
    \item[] Question: Does the paper describe the usage of LLMs if it is an important, original, or non-standard component of the core methods in this research? Note that if the LLM is used only for writing, editing, or formatting purposes and does \emph{not} impact the core methodology, scientific rigor, or originality of the research, declaration is not required.
    \item[] Answer: \answerNA{} 
    \item[] Justification:  Our work did not involve LLMs used in any novel, important or non standard way.

\end{enumerate}

\end{document}